\documentclass[10pt,twocolumn,letterpaper]{article}

\usepackage{cvpr}              

\usepackage{amsmath}

\usepackage{amssymb}
\usepackage{pifont}
\usepackage{color} 
\usepackage{soul} 
\newcommand{\cmark}{\ding{51}}%
\newcommand{\xmark}{\ding{55}}%
\usepackage{multirow} 

\definecolor{cvprblue}{rgb}{0.21,0.49,0.74}
\usepackage[pagebackref,breaklinks,colorlinks,allcolors=cvprblue]{hyperref}

\def\paperID{*****}
\def\confName{CVPR}
\def\confYear{2026}

\title{OVSegDT: Segmenting Transformer for Open-Vocabulary Object Goal Navigation}

\author{
Tatiana Zemskova$^{1,2}$   
\and 
Aleksei Staroverov$^{1,2,3}$
\and 
Dmitry Yudin$^{1,2}$
\and 
Aleksandr Panov$^{1,2}$   
\and
\\
$^1$Cognitive AI Systems Lab \hspace{2em} 
$^2$MIRAI  \hspace{2em} 
$^3$NUST MISIS
}

\begin{document}
\maketitle

\begin{abstract}

Open-vocabulary Object Goal Navigation requires an embodied agent to reach objects described by free-form language, including categories never seen during training. Existing end-to-end policies tend to overfit small simulator datasets, achieving high success on training scenes but failing to generalize and often exhibiting unsafe behavior (frequent collisions).
In our work, we are the first to show that a high degree of generalization to unseen categories in the open-vocabulary object goal navigation task can be achieved with a lightweight transformer model (130M parameters) using only RGB input. We introduce the OVSegDT approach, which has three key features.
First, we add a goal binary mask encoder that grounds the textual goal and provides precise spatial cues. The second component is a proposed Entropy-Adaptive Loss Modulation (EALM) — a per-sample scheduler that continuously balances imitation and reinforcement signals according to policy entropy, eliminating brittle manual phase switches. EALM reduces the sample complexity of training by 33\% and cuts the collision count by 10\% compared to the baseline.
The final component improves the agent’s navigation quality even under noisy predicted segmentation by combining an auxiliary segmentation loss with a reward function based on the area of the true goal mask during fine-tuning on predicted segmentation.
On HM3D-OVON, our model achieves performance on unseen categories comparable to that on seen ones and establishes state-of-the-art results (44.7\% SR, 20.6\% SPL on val unseen) without using depth, odometry, or large vision–language models. Code is available at \url{https://github.com/CognitiveAISystems/OVSegDT} .

\end{abstract}


\section{Introduction}

\begin{figure}[ht]
\begin{center}
\centerline{\includegraphics[width=1\columnwidth]{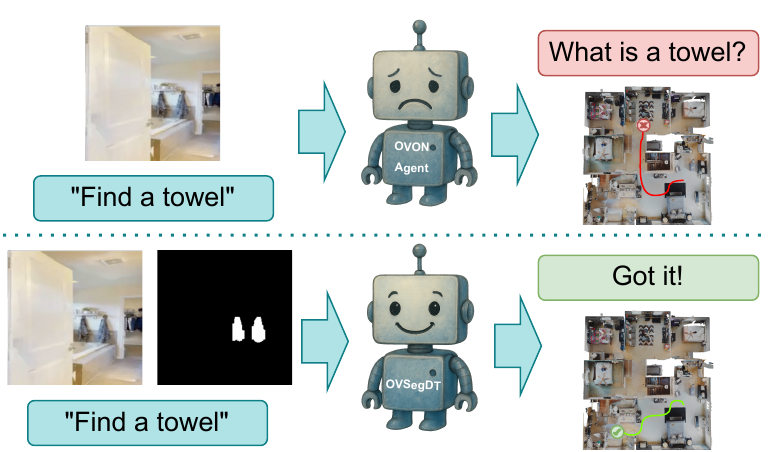}}
\caption{We demonstrate that effective generalization to unseen object categories during navigation is achievable using a lightweight transformer model OVSegDT that relies solely on RGB input and is trained in an end-to-end manner. Generalization is achieved through a target mask encoder and a training strategy that enhances both learning speed and navigation performance when using predicted segmentations from an external module.}
\label{fig:ga}
\end{center}
\end{figure}

Enabling robots to autonomously navigate toward a target object in an unfamiliar indoor environment is a critical capability for a wide range of real-world applications, including domestic assistance, search and rescue, and human-robot interaction. In this setting, the agent must perceive its surroundings, understand the task instruction, and explore the environment effectively to locate and approach the specified object. This task, commonly referred to as Object Goal Navigation (ObjectNav), presents a number of challenges, particularly when the environment is previously unseen and the object is described using open-ended, natural language.

Early work~\cite{staroverov2020real, maksymets2021thda, ramrakhya2023pirlnav} on ObjectNav typically framed the problem as navigation to one of a fixed set of object categories (e.g., "chair", "table"). While this formulation simplifies the task, it falls short of reflecting the complexity and flexibility of real-world scenarios. A more realistic setup is one where the agent can navigate to an arbitrary object category specified by a natural language query, including categories that were not seen during training~\cite{yokoyama2024hm3d}. This open-ended formulation better aligns with how humans describe goals and greatly increases the generalization ability of the agent.

The ObjectNav task involves exploring an environment, recognizing a target object, and navigating to it within a specified distance.
Existing solutions handle this pipeline either through modular approaches~\cite{staroverov2020real, yin2024sg, cai2024bridging}, which separate the system into dedicated skills (e.g., environment exploration, goal recognition, and goal reaching), or via end-to-end methods~\cite{maksymets2021thda, zhang2024uni, yokoyama2024hm3d}, where a single model implicitly learns to switch between different stages of the pipeline.

Modular map-based approaches excel in open-vocabulary object goal navigation~\cite{yokoyama2024vlfm, yin2024sg, chang2023goat} by combining language models for decision-making with open-vocabulary object detectors to build semantic maps. Detected objects are projected onto a spatial map, reducing the task to navigating to a location via deterministic~\cite{sethian1999fast} or learned policies~\cite{wijmansdd}.
However, map-based pipelines are sensitive to semantic map quality, with detection or pose errors compounding and reducing real-world performance. End-to-end mapless approaches~\cite{cai2024bridging, zhang2024uni, yokoyama2024hm3d} operate directly on raw sensor inputs without explicit intermediate representations. This reduces error accumulation, enables robust perception-language-control integration, and requires minimal sensors, making it ideal for resource-constrained real-world deployment.

A central challenge for end-to-end, mapless ObjectNav in open-vocabulary settings is the availability of photo-realistic training data. Even the largest publicly available simulator dataset, HM3DSem~\cite{yadav2023habitat}, comprises only 216 annotated scenes, which is orders of magnitude smaller than the corpora used to pretrain modern vision–language foundation models. Consequently, learned policies fail to generalize to novel object categories~\cite{yokoyama2024hm3d}.
To address this, hybrid mapless methods decouple exploration from goal-directed navigation. An exploration policy first locates candidate objects, then control is handed to a specialized navigation skill. This switch can be effected via binary object masks~\cite{cai2024bridging} or by estimating 3D goal coordinates from depth and odometry measurements~\cite{yokoyama2024hm3d}. These approaches combine the sample efficiency of modular design with the adaptability of end-to-end learning, though they often require additional sensors.

In our work, we are the first to demonstrate that effective generalization to unseen object categories during navigation is achievable using a lightweight transformer model (130M parameters) that relies solely on RGB input and is trained in an end-to-end manner. Generalization to new object categories is enabled by the encoder of the target object's binary mask (see~\cref{fig:ga}). We equip the proposed architecture with our training method designed to improve both learning speed and navigation quality when using predicted segmentation from an external segmentation module.

End-to-end ObjectNav approaches typically rely on a fragile two‐stage learning pipeline: an initial behavior cloning phase (e.g., DAgger) to bootstrap exploration, followed by a hard switch to Proximal Policy Optimization (PPO) for refinement.  The transition point between these objectives is chosen heuristically, often leading to catastrophic performance drops when the learning signals conflict~\cite{yokoyama2024hm3d}.  To overcome this, we introduce Entropy-Adaptive Loss Modulation (EALM), which dynamically mixes imitation and reinforcement signals at the sample level.  By deriving the PPO loss weight from policy entropy in each mini‐batch, EALM smoothly shifts emphasis from imitation to on-policy learning without manual scheduling, enabling stable, single‐stage training.

At the initial training stage, we use ground-truth target segmentation to accelerate convergence. However, in a realistic setup, ground-truth segmentation is unavailable. To obtain the target binary mask, we propose using a pretrained open-vocabulary segmentation model, which facilitates sim-to-real transfer without overfitting to simulated scenes for perception.
To reduce the gap between navigation performance with predicted segmentation compared to ground-truth segmentation, we introduce a combination of an auxiliary segmentation loss function and a reward based on the area of the ground-truth segmentation mask. Together, these components enable the OVSegDT model to adapt to predicted segmentation during fine-tuning, including for unseen target categories.

Thanks to the proposed architecture and training method, our approach, OVSegDT, achieves state-of-the-art results on the open-vocabulary object goal navigation benchmark HM3D-OVON~\cite{yokoyama2024hm3d} in a photorealistic environment, using only an RGB sensor and without relying on any additional sensors or large vision-language models.

\textbf{To summarize}, our main contributions are as follows:

\begin{itemize}

\item OVSegDT, a transformer-based architecture for open-vocabulary navigation, that incorporates a semantic encoder for binary target masks into the policy’s observation space. This goal mask encoder enables robust navigation to both seen and unseen object categories, even with noisy predicted segmentation.
\item Entropy‐Adaptive Loss Modulation (EALM), an automatic, entropy-gated DAgger–PPO switcher that uses policy entropy as a normalized uncertainty measure to interpolate between imitation and reinforcement signals, eliminating manual phase scheduling.
\item Combination of an auxiliary segmentation loss and a reward based on ground-truth mask area that together help OVSegDT adapt to predicted segmentation during fine-tuning and increase the generalization to unseen object categories.

\end{itemize}

\section{Related works}

The task of navigating to a target object described in natural language has gained attention as an extension of classic Object Goal Navigation (ObjectNav). Unlike a closed-vocabulary setting where the target object belongs to a predefined set of categories, open-vocabulary navigation requires agents to generalize to novel unseen object categories at inference time. In this task, the category names are specified as text, so open-vocabulary object goal navigation falls under the domain of Vision-Language Navigation (VLN).

\textbf{Map-based open-vocabulary ObjectGoal navigation.} Several works incorporate explicit mapping and pre-trained vision-language models to support open-vocabulary understanding. Early works that leverage large language models and explicit mapping, such as NavGPT~\cite{zhou2024navgpt} and MapGPT~\cite{chen2024mapgpt}, focused on the Room-to-Room navigation task~\cite{anderson2018vision}. This task requires following language instructions to navigate waypoints, while open-vocabulary object goal navigation involves autonomous search in unknown environments, making it more realistic for real-world robots. Considering methods for open-vocabulary object goal navigation, it is worth noting the VLFMs (Vision-Language Frontier Maps)~\cite{yokoyama2024vlfm} method, proposing constructing semantic frontier maps that leverage vision-language models to guide navigation without requiring task-specific training. In parallel, ESC~\cite{zhou2023esc} employs soft commonsense constraints during exploration to improve sample efficiency and reduce semantic ambiguity in object-goal navigation. Similarly, SG-Nav~\cite{yin2024sg} introduces dynamic 3D scene graph prompting, allowing large language models to reason over structured spatial representations and navigate to unseen objects. TANGO~\cite{ziliotto2025tango} proposes a neuro-symbolic, LLM-based framework for embodied AI that leverages primitive modules and extends prior exploration policies to multi-goal settings via memory map mechanisms. While effective, these approaches depend heavily on the quality of the map and robot localization, and errors in either can significantly degrade performance. At the same time, our method does not rely on building a map of the surrounding environment, and thus is independent of the quality of depth and odometry sensors.

\textbf{Mapless open-vocabulary ObjectGoal navigation.} In contrast, mapless methods for open-vocabulary ObjectGoal attempt to learn a policy that directly maps observations and language instructions to actions without explicit mapping. These works include, in particular, end-to-end approaches that map observations directly to action predictions. One such method is DagRL~\cite{yokoyama2024hm3d}, which shows that integrating pretrained visual backbones and conditioning the policy on object goal text embeddings enables generalization to unseen categories. PSL~\cite{sun2024prioritized} introduces a training strategy that prioritizes semantic learning, improving zero-shot navigation. However, end-to-end models are typically trained on millions of observations collected in a limited number of environments, such as HM3DSem~\cite{yadav2023habitat}, ProcTHOR~\cite{deitke2022}, and AI2Thor~\cite{kolve2017ai2}, which can hinder generalization to novel goals. To address this, UniNavid~\cite{zhang2024uni} leverages pretrained language models, while methods like PoliFormer~\cite{zeng2025poliformer} and DagRL-OD~\cite{yokoyama2024hm3d} incorporate visual cues from open-vocabulary object detectors. In our experiments, we use semantic masks of the target to provide precise location and boundary cues. PixelNav~\cite{cai2024bridging} also uses such masks but converts them into pixel goals and relies on additional LVLM and LLM modules for global planning. In contrast, we demonstrate that a lightweight transformer model (130M parameters) can efficiently explore environments and leverage open-vocabulary segmentation masks for effective open-vocabulary navigation.

\textbf{Learning Auxiliary Tasks for Navigation.}
To improve policy robustness in high-dimensional, multimodal settings, prior work leverages multi-task learning with auxiliary objectives that enhance representation quality and strengthen visual-language grounding for navigation.

For transformer-based navigation models, there are two common types of auxiliary tasks. The first type includes linguistic tasks such as visual question answering (UniNavid~\cite{zhang2024uni}, NaVILA~\cite{cheng2024navila}) and reasoning~\cite{wang2025think}. However, such tasks are typically effective for large transformer models, whereas we use a lightweight state encoder with only 34M parameters. The second type involves purely visual tasks, where the vision-language-action model is pre-trained to generate visual outputs from textual instructions and then fine-tuned to predict actions. Examples include GR-1~\cite{wu2024unleashing} and 3D-VLA~\cite{zhen20243d}, where the model learns to predict the goal’s final state, represented by an RGB image and a depth map. PixelNav~\cite{cai2024bridging} additionally uses tracking and step-distance to the goal as auxiliary losses. In contrast, we introduce a training method that enables effective adaptation of navigation method to predicted segmentation through a novel combination of the segmentation loss function and ground-truth semantic reward.

In previous works such as A2CAT-VN~\cite{kulhanek2019vision} and MTU3D~\cite{zhu2025move}, segmentation was used as an auxiliary task to help form better representations of the current observation. In contrast, we use segmentation as an auxiliary task for constructing the current state representation conditioned on past observations - that is, our transformer model jointly solves both navigation and semantic segmentation tasks. In this sense, our work is related to the Unified-IO2~\cite{lu2024unified} method, which builds a universal instruction-following model capable of both action prediction and image segmentation. 
Unlike this 1B-parameter model, OVSegDT is lightweight and incorporates observation history for navigation tasks, whereas Unified-IO2 is limited to manipulation tasks. 
We demonstrate that combining an auxiliary segmentation loss with the input goal mask and ground-truth semantic reward improves navigation performance toward text-specified goals, both for object categories seen during training and for unseen ones. This enables a lightweight generalizable approach to open-vocabulary object navigation using only an RGB sensor.

\section{Task Setup}

We formulate the Open-Vocabulary Object Navigation (OVON) task as a partially observable Markov decision process (POMDP) $ \langle \mathcal{S}, \mathcal{A}, P, R, \rho_0, \gamma \rangle$ with state space $\mathcal{S}$, discrete action space $\mathcal{A}$, transition distribution $P$, reward function $R$, initial state distribution $\rho_0$, and discount factor $\gamma$.

At every timestep $t$, the agent receives the observation
$o_t = \bigl(I_t,\; g,\; a_{t-1},\; M_{t}\bigr),$
where $I_t \in \mathbb{R}^{H\times W\times 3}$ is the current RGB frame, $g$ is the natural--language goal description, 
$a_{t-1}\!\in\!\{0,\dots,5\}$ is the previous discrete action, 
and $M_{t}\in\{0,1\}^{H\times W}$ is the binary segmentation mask for the target category at the current step.

The action set consists of six high--level motor commands: \texttt{stop}, \texttt{forward} (0.25\,m), \texttt{turn\_left} ($30^{\circ}$) and \texttt{turn\_right} ($30^{\circ}$), \texttt{look\_up} ($30^{\circ}$), and \texttt{look\_down} ($30^{\circ}$). An episode terminates when either \texttt{stop} is issued or the horizon of 500 steps is reached. 

We adopt the standard ObjectNav metrics -- Success Rate (SR), Success weighted by Path Length (SPL), and SoftSPL -- following \citet{yokoyama2024hm3d}.

\section{Method}


 In this section, we first outline the architectural features of our method, then present the training objectives used to accelerate OVSegDT convergence in the interactive simulator, and finally discuss adapting the method to real-world settings without simulator ground-truth segmentation.

\label{sec:preliminaries}
\paragraph{Notation.} The transformer policy is denoted by $\pi_\theta(a\mid o)$ with parameters $\theta$ and value head $V_\theta(o)$. Advantage estimates~\cite{schulman2015high} are $\hat A_t$ and returns $\hat R_t$.

\paragraph{Proximal Policy Optimisation (PPO).}
For an on--policy batch $\{(o_t,a_t,\hat A_t,\hat R_t)\}$ we minimize the clipped surrogate
\begin{equation}
\mathcal{L}_{\text{PPO}}(\theta)=\mathbb{E}_t\Bigl[-\min( r_t\hat A_t,\; \operatorname{clip}(r_t,1-\varepsilon,1+\varepsilon)\hat A_t)\Bigr],
\end{equation}
\begin{equation}
r_t=\frac{\pi_\theta(a_t\mid o_t)}{\pi_{\theta_{\text{old}}}(a_t\mid o_t)}.
\end{equation}
A value term $\mathcal{L}_V(\theta)=\tfrac12\lVert V_\theta(s_t)-\hat R_t\rVert_2^2$ and entropy bonus $H_t$ are added following standard practice~\cite{schulman2017proximal} (see~\cref{eq:total}).

\paragraph{Dataset Aggregation (DAgger).}
We follow the DAgger implementation from~\cite{yokoyama2024hm3d}. Given teacher actions $a_t^E$ from a classical A* planner~\cite{hart1968formal} that uses a ground-truth map and GPS sensor with a frontier-based exploration strategy for goal object searching~\cite{santosh2008autonomous}, behavior cloning minimizes
\begin{equation}
\mathcal{L}_{\text{BC}}(\theta)=\mathbb{E}_t[-\log \pi_\theta(a_t^E\mid o_t)].
\end{equation}
Importantly, our implementation also updates the critic during behavior cloning, enabling a seamless transition to RL.

\begin{figure}[ht]
\begin{center}
\centerline{\includegraphics[width=1.0\columnwidth]{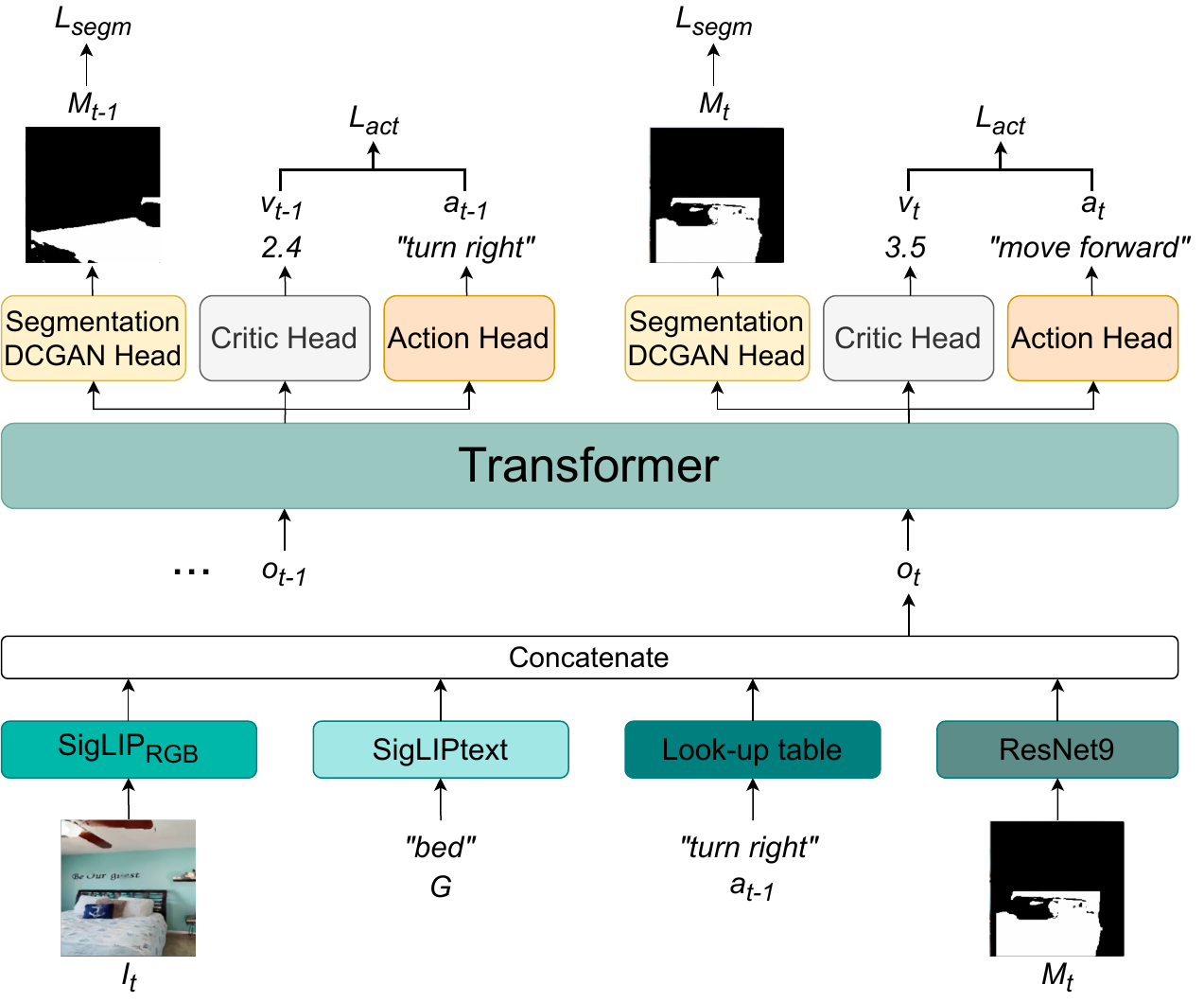}}
\caption{OVSegDT model encodes the current visual observation  $I_t$, the goal object category $G$, the previous action $a_{t-1}$ and binary mask of the target object (goal mask) $M_t$ to form observation embedding $o_t$. At each timestep, a transformer receives the embedding sequence from the previous 100 steps. The action head is used to sample the action $a_t$. During training, a segmentation DCGAN head generates a binary target mask $m_t$ for the auxiliary segmentation loss function, and a critic head is employed to predict the value $v_t$ of the current state.
}
\label{fig:model_scheme}
\end{center}
\end{figure}

\subsection{OVSegDT Architecture}

Following the state-of-the-art architecture for open-vocabulary navigation proposed by the authors of the HM3D-OVON~\cite{yokoyama2024hm3d} benchmark, we use a frozen SigLIP~\cite{zhai2023sigmoid} encoder for RGB images and text, as these encoders have demonstrated strong performance on the ObjectNav task~\cite{ehsani2023imitating}. To encode the observation at time step $t$, the image encoder $\text{SigLIP}_{\text{RGB}}$ produces a 768-dimensional embedding $i_t = \text{SigLIP}_{\text{RGB}}(I_t)$ for the visual input $I_t$. Similarly, the text encoder $\text{SigLIP}_{\text{text}}$ generates a 768-dimensional embedding $g_t = \text{SigLIP}_{\text{text}}(G)$ for the object category $G$ provided as text.

We also use a learnable 32-dimensional embedding for the previous discrete action $p_t = \phi_a(a_{t-1})$. To encode the binary target mask $M_t$, we use a simple, lightweight ResNet9 architecture~\cite{he2016deep}, which has proven efficient in our experiments. This results in a 128-dimensional goal mask embedding $m_t = \text{ResNet9}(M_t)$. The resulting embeddings are concatenated to form the observation embedding $o_t = [ i_t, g_t, p_t, m_t ]$.

A sequence of the most recent 100 observation embeddings $[ o_{t-99}, \ldots, o_t ]$ is passed to a decoder-only transformer model $\pi_{\theta}$~\cite{vaswani2017attention}. We adopt the same transformer architecture found effective in the original OVON method: 4 layers, 8 heads, hidden size 512, and a maximum context length of 100. At each timestep, $\pi_{\theta}$ predicts a feature vector, which is passed to decoding heads.

The action head is implemented as a linear layer that predicts a categorical distribution from which the action $a_t$ is sampled. 
During training, we also use two additional heads: a critic head, implemented as a linear layer, and a segmentation head, implemented as a DCGAN~\cite{radford2015unsupervised} model that reconstructs the current binary target mask $M_t$ from the feature vector predicted by the transformer $\pi_{\theta}$.

\subsection{Training Objectives}

In this section, we describe the training objectives used to train our OVSegDT model. Our primary goal was to develop a unified single-stage training pipeline to obtain a general-purpose navigation model capable of using binary target masks as visual cues for navigating to object categories that were not present in the training dataset.

\textbf{Reward function.}
\label{sec:semantic_reward}
To train OVSegDT, we use a combination of the standard reward for the end-to-end ObjectNav task and the semantic reward proposed in~\cite{staroverov2025semantic}. The agent receives a sparse reward of $+2.5$ for successfully completing an episode and a small negative reward for collisions. Let $d_t$ represent the geodesic distance from the agent to the target object at time $t$, and let $b_t$ indicate the area (in pixels) of the ground-truth binary mask of the target object in the current frame. For each step we define the semantic reward as the following:

\begin{equation}
 R^{sem}_t= \lambda_{sem}(\mathrm{clip}(d_{t-1}-d_t)+\mathrm{clip}(b_t-b_{t-1})),
\end{equation}

where the first term encourages progress toward the target, while the second term rewards an increased presence of the ground-truth target in the current observation as indicated by the segmentation mask. $\lambda_{sem}$ is set to $0.5$, and the $\mathrm{clip}$ function constrains both terms to the range $[-1, 1]$.

\subsubsection{Entropy-Adaptive Loss Modulation (EALM)}
\label{sec:ealm}

Open-vocabulary navigation requires a policy to decide 
how strongly it should rely on imitation signals versus exploration-driven reinforcement learning.  
EALM solves this by continuously blending Behavior Cloning (BC) and Proximal Policy Optimization (PPO) objectives according to the policy's uncertainty, quantified by its action entropy. 

\textbf{Entropy tracking.}
For every mini--batch we compute the Shannon entropy of the action distribution
\begin{equation}
  H_t \;=\; -\!\!\sum_{a\in\mathcal{A}} \pi_{\theta}(a\mid o_t)\, \log \pi_{\theta}(a\mid o_t),
\end{equation}
and maintain an exponential moving average (EMA):
  $\hat H_t \;=\; \alpha\, \hat H_{t-1} \; + \; (1-\alpha)\, H_t, \quad 0<\alpha<1$.
We adopt $\alpha=0.95$ throughout. This exponential smoothing filters out high-frequency entropy fluctuations while preserving the overall learning trend, providing a stable signal for loss weighting.

\textbf{Adaptive gate.}
Given two entropy bounds $H_{\text{low}} < H_{\text{high}}$ (we use $0.35$ and $0.75$ in practice),
the EMA is mapped to a mixing coefficient
\begin{equation}
  \lambda_t \;=\;
  \mathrm{clip}\!\left( \frac{H_{\text{high}} - \hat H_t}{H_{\text{high}}-H_{\text{low}}},\, 0,\, 1 \right).
\end{equation}
Low entropy (confident policy) therefore yields $\lambda_t\!\approx\!1$ (pure PPO), whereas very high entropy favors BC.
We further define  $p_{\text{PPO}}\!=\!\lambda_t$, $p_{\text{BC}}\!=\!1-\lambda_t$.

The choice of entropy bounds is derived from the action space structure. For $|\mathcal{A}|=6$ discrete actions, the maximum entropy is $H_{\max}=\log 6 \approx 1.79$ nats. We set $H_{\text{high}} = 0.42 \cdot H_{\max} \approx 0.75$ (corresponding to $\approx$80\% top-action probability) and $H_{\text{low}} = 0.20 \cdot H_{\max} \approx 0.35$ (corresponding to $\approx$95\% top-action probability). This normalization generalizes across different action spaces: as a rule of thumb, $H_{\text{high}} \in [0.4, 0.5] \cdot \log|\mathcal{A}|$ and $H_{\text{low}} \in [0.15, 0.25] \cdot \log|\mathcal{A}|$ provide robust performance.

\textbf{EALM objective.}
With $\mathcal{L}_{\text{PPO}}$ the usual clipped surrogate and $\mathcal{L}_{\text{BC}}=-\log\pi_{\theta}(a_t^{E}\mid o_t)$ the imitation loss, 
the policy term becomes
\begin{equation}
  \mathcal{L}_{\text{EALM}}(\theta) \;=\; p_{\text{PPO}}\,\mathcal{L}_{\text{PPO}}(\theta) 
  \; + \; p_{\text{BC}}\,\mathcal{L}_{\text{BC}}(\theta).
\end{equation}

EALM therefore provides a smooth, automatic transition from demonstration-driven learning to pure reinforcement learning, eliminating the need for manual phase scheduling.

\subsubsection{Auxiliary Semantic Segmentation Loss}
\label{sec:aux_segm_loss}
To encode the observation, we use a single vector obtained by concatenating different types of information extracted from the environment. This helps reduce the time required to predict the next action and enables the model to retain a long observation history, which is particularly important for efficient learning in navigation tasks. However, when working with such a concatenated observation vector, the transformer model must learn how to effectively utilize the information contained in different parts of the observation.
In our experiments, we show that using an auxiliary semantic segmentation loss allows the transformer to learn effective navigation representations more quickly during the early stages of training, where the behavior cloning term dominates the navigation training objective.
As the loss function for semantic segmentation, we use the sum of the Dice loss and binary cross-entropy loss, which are commonly used as training objectives for binary mask segmentation (e.g., in~\cite{cheng2022masked}):

\begin{equation}
\mathcal{L}_{\text{seg}}(\theta) = \mathcal{L}_{\text{Dice}}(\theta) + \mathcal{L}_{\text{CE}}(\theta).
\end{equation}

\subsubsection{Total Training Loss} Finally, the overall optimization target for a mini--batch is
\begin{equation}
\label{eq:total}
  \mathcal{L}_{\text{total}}(\theta) 
  = c_v\,\mathcal{L}_V(\theta) + L_{\text{EALM}}(\theta) - \beta H_t + \mathcal{L}_{\text{seg}}(\theta),
\end{equation}
where $ \mathcal{L}_V$ is the value regression term. 
All coefficients are kept task--invariant across experiments.

\subsection{Adaptation to Predicted Segmentation}

Our experiments show that using a binary ground-truth mask of the target not only accelerates the model’s training process but also significantly improves the adaptability of the approach to new environments and target categories. However, in real-world scenarios, ground-truth segmentation is not available, and the quality of navigation heavily depends on the performance of the pre-trained segmentation model used. Given the universal nature of the task and the availability of modern open-vocabulary segmentation models, we aim to leverage a pre-trained segmentation model, as this allows us to minimize the sim-to-real gap for segmentation.

In our experiments, we use a pre-trained YOLOE~\cite{wang2025yoloe} model for open-vocabulary segmentation -- a state-of-the-art model with real-time inference speed. When using this model out of the box, we observe a significant drop in navigation performance compared to using ground-truth segmentation. To mitigate this drop, we apply two strategies.

First, YOLOE struggles to recognize masks of objects that are distant or heavily occluded by other objects. To address this issue, we fine-tune our OVSegDT model using pure PPO loss 
using the predicted masks from the YOLOE model at each step. During such fine-tuning, it is important to maintain the OVSegDT ability to generalize to unseen object categories, since the agent may receive a reward for successfully completing an episode even when the target segmentation mask is missing. In our case, generalization to unseen object categories is achieved by using a combination of the semantic reward (\cref{sec:semantic_reward}) and auxiliary segmentation loss (\cref{sec:aux_segm_loss}), both during training on ground-truth segmentation and during fine-tuning on predicted segmentation. It is precisely their combination that enables OVSegDT to effectively adapt to predicted segmentation (see \cref{sec:ablation_segm_training}).

Second, we found that different object categories pose varying levels of difficulty for the segmentation model. For example, accurately recognizing a "dishwasher" requires discarding all predicted masks with confidence below 0.4, whereas detecting objects such as "book" or "flowerpot" requires keeping all masks with confidence no lower than 0.01. To address this, we calibrate the detector's confidence thresholds for different object categories in the model's navigation vocabulary, which significantly improves navigation quality using predicted segmentation (see \cref{tab:obs_seg_quality}).
Finally, we remove semantically redundant categories (e.g., "rug" and "carpet") from the segmentation vocabulary, keeping only one of each such set.  In practice, a confidence threshold of 0.3 is generally robust for segmentation and therefore navigation to unseen categories. The described adaptation of the segmentation model to the specific scenario of robot deployment is a significantly less expensive operation than retraining the navigation model, which takes tens of millions of steps.
\begin{table*}[t]
\centering
\caption{Open-Vocabulary Object goal navigation. Comparison on HM3D-OVON~\cite{yokoyama2024hm3d}.}
\scriptsize
\begin{tabular}{lllcccccc}

\hline
\textbf{Method} & \textbf{Depth} & \textbf{Odometry} & \multicolumn{2}{c}{\textbf{Val Seen}} & \multicolumn{2}{c}{\textbf{Val Seen Synonyms}} & \multicolumn{2}{c}{\textbf{Val Unseen}} \\
 &  &  & \textbf{SR$\uparrow$} & \textbf{SPL$\uparrow$} & \textbf{SR$\uparrow$} & \textbf{SPL$\uparrow$} & \textbf{SR$\uparrow$} & \textbf{SPL$\uparrow$} \\

\hline
BC~\cite{yokoyama2024hm3d}     & \xmark & \xmark & $ 11.1 \pm 0.1 $ & $4.5 \pm 0.1 $ & $9.9 \pm 0.4 $ & $3.8 \pm 0.1 $ & $5.4 \pm 0.1 $ & $1.9 \pm 0.2$\\
DAgger~\cite{yokoyama2024hm3d} & \xmark & \xmark & $ 18.1 \pm 0.4$  & $ 9.4 \pm 0.3$  & $ 15.0 \pm 0.4$ & $ 7.4 \pm 0.3$  & $ 10.2 \pm 0.5$  & $ 4.7 \pm 0.3$ \\
RL~\cite{yokoyama2024hm3d}     & \xmark & \xmark & $ 39.2 \pm 0.4$ & $ 18.7 \pm 0.2$ & $ 27.8 \pm 0.1$ & $ 11.7 \pm 0.2 $  & $ 18.6 \pm 0.3$  & $ 7.5 \pm 0.2$  \\
BCRL~\cite{yokoyama2024hm3d}   & \xmark & \xmark & $ 20.2 \pm 0.6$  & $ 8.2 \pm 0.4$ & $ 15.2 \pm 0.1$  & $ 5.3 \pm 0.1$  & $ 8.0 \pm 0.2$ & $ 2.8 \pm 0.1$ \\
DAgRL~\cite{yokoyama2024hm3d}  & \xmark & \xmark & $ 41.3\pm 0.3$ & $ 21.2 \pm 0.3$ & $ 29.4 \pm 0.3 $ & $ 14.4 \pm 0.1$  & $ 18.3 \pm 0.3$ & $ 7.9 \pm 0.1$ \\
Uni-NaVid~\cite{zhang2024uni} & \xmark & \xmark & $ 41.3 $  & $ \underline{21.1} $  & $ \underline{43.9} $ & $ \mathbf{21.8} $ & $ 39.5 $ & $ \underline{19.8} $ \\
\hline
VLFM~\cite{yokoyama2024vlfm} & \cmark & \cmark & 35.2 & 18.6 & 32.4 & 17.3 & 35.2 & 19.6 \\
DAgRL+OD~\cite{yokoyama2024hm3d} & \cmark & \cmark & $ 38.5 \pm 0.4$ & $ 21.1 \pm 0.4$  & $ 39.0 \pm 0.7$ & $ \underline{21.4} \pm 0.5$ & $ 37.1 \pm 0.2 $ & $\underline{19.8}  \pm 0.3 $\\

TANGO~\cite{ziliotto2025tango} & \cmark & \cmark & - & - & - & - & $ 35.5  \pm 0.3$ & $ 19.5  \pm 0.3$ \\
MTU3D~\cite{zhu2025move} & \cmark & \cmark  & $\mathbf{55.0}$ & $\mathbf{23.6 }$& $\mathbf{45.0}$ & 14.7 & \underline{40.8} & 12.1 \\
\hline
OVSegDT & \xmark & \xmark & $ \underline{43.6} \pm 0.4$ & $20.1 \pm 0.2 $ & $ 40.1 \pm 0.4 $ & $ 17.9 \pm 0.1 $ & $ \mathbf{44.7} \pm 0.4 $ & $ \mathbf{20.6}\pm 0.2 $  \\
\hline

\end{tabular}

\label{tab:results_ovon_main}
\end{table*}

\section{Results}
\label{sec:results}

\textbf{Training details.} We train all policies using ground-truth segmentation for 200M steps, as incremental improvements diminished beyond this point. We perform additional fine-tuning on the predicted segmentation for 15M steps.
All experiments were conducted across 40 environments distributed across 2 NVIDIA A100 GPUs utilizing Variable Experience Rollout~\cite{wijmans2022ver}. We provide the list of used hyper-parameters and navigation reward details in the Sup. Mat. (Sec. A).

\textbf{Experimental setup.} We evaluate the effectiveness of different methods using three metrics: SR, SPL, and the average number of collisions per episode. For methods that sample actions from distributions during navigation, we average the results across three seeds and report the mean and standard deviation. Details of the statistical analysis are provided in the Sup. Mat. (Sec. F).

\subsection{Comparison with the State-of-the-Art Methods} 
\begin{figure}[ht]
\begin{center}
\centerline{\includegraphics[width=1.0\columnwidth]{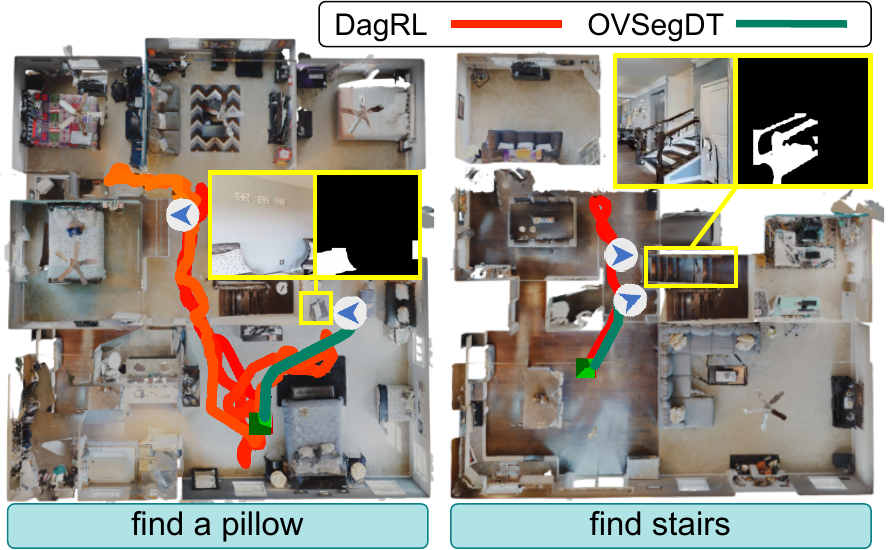}}
\caption{Qualitative comparison of navigation trajectories between our method OVSegDT and the baseline DagRL on categories from the \textit{val unseen} split of the HM3D-OVON benchmark.
\textbf{Left}: The DagRL model fails to recognize the target object \textit{a pillow} and thus ends the episode unsuccessfully, while OVSegDT successfully reaches the goal thanks to the binary goal mask.
\textbf{Right}: The use of the binary goal mask enables OVSegDT to search for the target object \textit{stairs} more efficiently compared to DagRL, which takes extra steps in the environment and passes by the goal at the beginning of the episode.
}
\label{fig:qual_examples}
\end{center}
\end{figure}

We compare our method to state-of-the-art approaches on HM3D-OVON~\cite{yokoyama2024hm3d}, a benchmark tailored for open-vocabulary object-goal navigation. In these experiments, we use the pretrained YOLOE~\cite{wang2025yoloe} model as the source of goal segmentation.~\cref{tab:results_ovon_main} shows the navigation performance of our approach compared to existing methods for open-vocabulary navigation. Notably, our method demonstrates improvements in both the success rate and the efficiency in terms of path length (SPL metric) over baseline RGB-only approaches that do not leverage large language models. We provide a trajectory analysis of our method compared to the baseline DagRL, which does not use segmentation masks as input, in~\cref{fig:qual_examples}. These examples illustrate that, during inference, the binary goal mask helps the agent recognize the target object and switch from exploration to goal-directed navigation, while also reducing the distance traveled to reach the target. It is worth noting that our approach outperforms large VLN models such as Uni-NaVid~\cite{zhang2024uni} and TANGO~\cite{ziliotto2025tango}, as well as methods that rely on additional sensors such as depth and odometry for navigation, including DAgRL+OD~\cite{yokoyama2024hm3d} and MTU3D~\cite{zhu2025move} on the val unseen split of HM3D-OVON. Furthermore, the binary goal mask input makes our approach highly generalizable, as the navigation performance on both val seen and val unseen categories is nearly identical.

\subsection{Ablation Studies}

We analyze the effectiveness of individual components of our approach through a series of ablation experiments.

\begin{table*}[t]
\centering
\caption{Quantitative comparison of switching strategies on HM3D-OVON benchmark. Besides classic navigation metrics (SR and SPL) we report mean collisions – a safety metric strongly correlated with generalisation (lower is better; bold indicates best). }
\tiny
\begin{tabular}{lccccccc}
\hline
\textbf{Method} & \textbf{Training steps} &  \multicolumn{3}{c}{\textbf{Val seen}} & \multicolumn{3}{c}{\textbf{Val unseen}} \\
 & & \textbf{SR$\uparrow$} & \textbf{SPL$\uparrow$} & \textbf{Collisions$\downarrow$} & \textbf{SR$\uparrow$} & \textbf{SPL$\uparrow$} & \textbf{Collisions$\downarrow$} \\
\hline
PPO                & 200M      & 7.3  & 3.2  & 6.0  & 0.5  & 0.3  & 1.3   \\
\hline
Dagger             & 200M      & 29.0 & 15.3 & 36.5 & 15.7 & 6.7  & 50.4  \\

Dagger+PPO         & 200M      & 29.2 & 16.0 & 31.1 & 14.3 & 6.5  & \underline{35.8}  \\
EarlySwitcher      & 200M      & 32.6 & 16.9 & \underline{25.6} & 16.4 & 7.5  & $\mathbf{27.7}$  \\
DagRL              & 300M      & $\underline{41.3} \pm 0.3 $ & $\underline{21.2} \pm 0.3 $ & 33.4 & $ \underline{18.3} \pm 0.3 $ & $ \underline{7.9} \pm 0.1 $  & 50.2  \\
\textbf{EALM (ours)}   & 200M & $ \mathbf{42.5} \pm 0.4 $ & $\mathbf{21.3} \pm 0.2 $ & $\mathbf{25.3} \pm 2.1 $ & $\mathbf{20.2} \pm 0.5 $ & $\mathbf{8.8} \pm 0.2 $ & $41.8 \pm 2.5 $ \\

\hline
\end{tabular}

\label{tab:results_ovon}
\end{table*}

\begin{table}[h]
\centering
\caption{Comparison of methods using different observation types and training objectives on Val Unseen split of HM3D-OVON benchmark. YOLOE segmentation masks are used as input.}
\tiny
\begin{tabular}{@{}p{0.8cm}p{0.4cm}p{0.8cm}p{0.1cm}p{0.1cm}p{1cm}p{1cm}p{1cm}@{}}
\hline
\textbf{Method} & \textbf{EALM } & \textbf{Goal mask} & \textbf{$r^{\text{sem}}$} & \textbf{$\mathcal{L}_{\text{seg}}$} & \textbf{SR$\uparrow$} & \textbf{SPL$\uparrow$} & \textbf{Collisions$\downarrow$}\\

\hline
DagRL & \xmark & \xmark & \xmark & \xmark & $ 18.3 \pm 0.3 $ & $7.9 \pm 0.1$ & $50.2$ \\
OVSegDT  & \cmark & \xmark & \xmark  & \xmark & $20.2 \pm 0.6 $ & $8.8 \pm 0.3 $ & $41.8 \pm 2.5 $ \\
OVSegDT  & \cmark & \cmark & \xmark  & \cmark & $26.7 \pm 0.5 $& $13.2 \pm 0.2 $ & $\mathbf{29.4} \pm 3.1$ \\
OVSegDT  & \cmark & \cmark & \cmark  & \xmark & $\underline{36.9} \pm 0.7 $ & $\underline{16.5} \pm 0.1 $ & $49.3\pm 0.4$ \\
OVSegDT & \cmark & \cmark & \cmark &\cmark & $\mathbf{44.7} \pm 0.4 $ & $\mathbf{20.6} \pm 0.2$ & $\underline{45.4} \pm 0.6 $ \\
\hline
\end{tabular}
\label{tab:obs_seg_results}
\end{table}

\begin{table}[h]
\centering
\caption{Analysis of adaptation strategies to predicted segmentation for OVSegDT on the HM3D-OVON Val Unseen split.}
\tiny
\begin{tabular}{@{}p{1.2cm}p{1.1cm}@{}p{1.2cm}@{}p{1.2cm}@{}p{1.2cm}@{}p{2cm}@{}}
\hline
\textbf{Predicted mask finetune} & \textbf{Segmentation source} & \textbf{SR$\uparrow$ }& \textbf{SPL$\uparrow$} & \textbf{Collisions$\downarrow$} & \textbf{Inference time (msec, NVIDIA Jetson Orin)} \\

\hline
\xmark & GT & $69.3 \pm 0.2 $ & $39.1 \pm 0.2$ & $23.3 \pm 0.3 $ & $107 \pm 5$ \\
\hline
\xmark & YOLOE & $  28.8 \pm 0.4 $ & $ 14.0 \pm 0.2$ & $\mathbf{19.6} \pm 0.2$ &  $132 \pm 5$ \\
\xmark & YOLOE+calib. & $ \underline{35.6} \pm 0.4 $ & $\underline{16.2} \pm 0.3$ & $\underline{20.2} \pm 0.8$ &  $132 \pm 5$ \\
\cmark & YOLOE+calib. & $\mathbf{44.7} \pm 0.4 $ & $\mathbf{20.6} \pm 0.2$ & $45.4 \pm 0.6 $ &  $132 \pm 5$ \\
\hline
\end{tabular}

\label{tab:obs_seg_quality}
\end{table}

\begin{table}[h]
\centering
\caption{Failure analysis on the HM3D-OVON Val Unseen split. 
}
\tiny

\begin{tabular}{@{}p{2.1cm}@{}p{1.3cm}@{}p{1.3cm}@{}p{1.3cm}@{}p{1.3cm}@{}p{1cm}@{}}
\hline
\textbf{Method} &
\textcolor{gray}{\textbf{OVSegDT (ours)}} &
\textbf{OVSegDT (ours)} &
\textbf{OVSegDT (ours) }&
\textbf{OVSegDT (ours)} &
\textbf{DagRL}
\\
\hline
Segmentation source &
\textcolor{gray}{GT} &
YOLOE &
YOLOE+calib. &
YOLOE+calib. &
- \\
\hline
Predicted mask finetune &
\textcolor{gray}{\xmark} &
\xmark &
\xmark &
\cmark &
- \\

\hline
Exploration failures, \% & \textcolor{gray}{$12.3 \pm 0.1$} & $12.1 \pm 0.3$  & $12.2 \pm 0.2$ & $\mathbf{8.9} \pm 0.1$ & $\underline{12.0} \pm 0.1$ \\
\hline
Goal Recognition failures, \% & \textcolor{gray}{$5.5 \pm 0.4$} & $37.8 \pm 0.4$ & $\underline{34.0} \pm 0.5$ & $\mathbf{28.2} \pm 0.5$ & $45.7 \pm 0.3$ \\
\hspace{1em}Ignored goal object, \% & \textcolor{gray}{$5.5 \pm 0.4$} & $\underline{16.6} \pm 0.3$ & $19.5 \pm 0.6$ & $\mathbf{15.8} \pm 0.4$ & $33.0 \pm 0.2$ \\
\hspace{1em}Stopped at wrong object, \% & \textcolor{gray}{$0.0 \pm 0.0$} & $21.2 \pm 0.4$ & $\underline{14.5} \pm 0.4$ & $\mathbf{12.5} \pm 0.3$ & $12.8 \pm 0.5$ \\
\hline
Goal Reaching failures, \%  & \textcolor{gray}{$13.5 \pm 0.1$} & $21.3 \pm 0.5$ & $\mathbf{18.1} \pm 0.5$ & $\underline{18.2} \pm 0.1$ & $24.5 \pm 0.4$ \\
\hspace{1em}Stopped too far, \% & \textcolor{gray}{$10.3 \pm 0.1$} & $17.0 \pm 0.4$ & $\underline{13.2} \pm 0.2$ & $\mathbf{12.0} \pm 0.4$ & $15.9 \pm 0.6$ \\
\hspace{1em}Moved away, \% & \textcolor{gray}{$3.2 \pm 0.2$} & $\mathbf{4.2} \pm 0.4$ & $\underline{4.9} \pm 0.3$ & $6.2 \pm 0.3$ & $8.6 \pm 0.2$ \\
\hline
Total Failures, \% & \textcolor{gray}{$31.3 \pm 0.4$} & $71.2 \pm 0.3$ & $\underline{64.3} \pm 0.4$ & $\mathbf{55.3} \pm 0.3$ & $82.2 \pm 0.2$ \\
\hline
\end{tabular}
\label{tab:failure_analysis}
\end{table}

\subsubsection{Comparison of DAgger-PPO Switching Methods}

To evaluate our proposed training paradigm, we conduct an extensive comparison of different learning approaches for open-vocabulary object navigation. We analyze four distinct strategies: (1) PPO, (2) DAgger (3) EarlySwitcher that smoothly switches from DAgger to PPO at steps 40M-60M, (4) DAgger+PPO combined loss, (5) DAgRL baseline~\cite{yokoyama2024hm3d}, (6) EALM, our proposed hybrid approach that jointly optimizes DAgger and PPO objectives.
All experiments use the same transformer-based architecture with SigLIP features for RGB encoding and text goal representation, following the same architecture as in \cite{yokoyama2024hm3d}.

~\cref{tab:results_ovon} confirms that naïve PPO almost stalls. Success Rate (SR) alone, however, does not capture \emph{how} the goal is reached.~\cref{tab:results_ovon} complements the analysis with the average \emph{collision count} per episode. This metric, part of the PPO reward, is a convenient safety proxy: fewer collisions imply that the agent actually understands how to navigate rather than memorizing trajectories. The absolute collision counts reported in~\cref{tab:results_ovon} should be understood in the context of episode length. For EALM on val unseen, the average successful episode length is approximately 450 steps, yielding a collision rate of $\sim$9\% (41.8 collisions / 450 steps). In contrast, pure PPO terminates early (mean episode length $<$15 steps), so its low collision count reflects failure rather than safe navigation.

DAgger attains near-perfect SR on the training scenes yet collides more than $50$ times per unseen episode, illustrating severe over-fitting. 
The combined DAgger+PPO baseline shows lower collisions, but as the PPO term could not completely dominate the DAgger loss, the policy does not show improvement in the SR.
EarlySwitcher demonstrates how delicate manual scheduling is. In summary, EALM proves crucial for producing policies that are \emph{both} successful \emph{and} safe. 
Training curves and experiments with varying lower entropy bounds are provided in the Sup. Mat. (Sec. B, Sec. C).

\subsubsection{Training Components} 
\label{sec:ablation_segm_training}
We analyze how each of our proposed components affects the performance of OVSegDT during training. In this experiment, we use YOLOE segmentation goal masks for the model variants that take a binary goal mask as input.~\cref{tab:obs_seg_results} shows that EALM improves overall navigation performance compared to the baseline training approach with a hard switch between DAgger and PPO objectives. Using the segmentation goal mask significantly boosts navigation performance on unseen categories. We conduct an analysis of the impact of the auxiliary segmentation loss function and the semantic reward, taking into account adaptation to predicted segmentation: fine-tuning on YOLOE predictions and model confidence calibration. The combination of the auxiliary segmentation loss and the semantic reward significantly improves both the success rate and SPL. The semantic reward increases the number of collisions; however, this increase is moderate compared to the improvement in navigation quality achieved with its use.

\subsubsection{Adaptation to Predicted Segmentation}
We analyze how segmentation quality affects the navigation performance of our method and evaluate the contribution of our proposed techniques for improving navigation when using predicted segmentation.~\cref{tab:obs_seg_quality} shows that using the out-of-the-box YOLOE segmentation~\cite{wang2025yoloe} leads to a significant drop in navigation performance for unseen object categories. Calibrating YOLOE confidence thresholds and vocabulary yields a gain of $+7$\% in SR and $+2.4\%$ in SPL thanks to more accurate object category recognition. Fine-tuning on YOLOE predicted masks yields an additional improvement of $+9.1$\% in SR and $+4.1\%$ in SPL but leads to an increase in the number of collisions. We believe that in the navigation task, the SR and SPL are the most important metrics; therefore, we retain this predicted-mask fine-tuning in subsequent experiments. Finding a more optimal integration of the segmentation and navigation reward functions for better optimization is left for future work. Overall, our method OVSegDT achieves a gain of $+26.4$\% in SR, $+12.7$\% in SPL, and a $10$\% reduction in the average number of collisions compared to the baseline approach DagRL on the val unseen split of HM3D-OVON. 
Predicted segmentation increases inference time on the NVIDIA Jetson Orin by only 23\%, keeping OVSegDT suitable for real-time onboard policy prediction. Sec. J in the Sup. Mat. shows demonstration experiments of OVSegDT on a real robot.


We analyze how predicted segmentation affects navigation error profiles in~\cref{tab:failure_analysis}. Most errors stem from goal recognition and goal reaching. Calibration reduces false positives and premature stopping, while our fine-tuning with auxiliary segmentation supervision and a semantic reward significantly decreases errors in goal recognition. Further analysis of the impact of segmentation errors can be found in the Sup. Mat. (Sec. H).

\section{Conclusion}

In this work, we presented OVSegDT, a novel approach for open-vocabulary object navigation  that achieves strong generalization to unseen object categories using a lightweight end-to-end transformer model with only RGB input from sensors. Our experiments demonstrate that our method significantly improves navigation performance, particularly on unseen object categories, by leveraging the synergy between segmentation and policy learning. Through the proposed Entropy-Adaptive Loss Modulation, we unify imitation and reinforcement learning into a stable single-stage training process, removing the need for heuristic phase switching. Furthermore, by incorporating an auxiliary segmentation loss and semantic reward, OVSegDT effectively adapts to predicted masks from open-vocabulary segmentation models.
Overall, our results on the HM3D-OVON benchmark demonstrate that lightweight architectures can achieve both robust generalization and state-of-the-art performance in open-vocabulary navigation tasks, paving the way toward scalable and perception-efficient embodied agents. The limitations of the current approach include its dependence on the quality of the segmentation model and the lack of the ability to follow free-form text instructions. These directions are the subject of future work.

\section*{Acknowledgments}
The study was supported by the Ministry of Economic Development of the Russian Federation (agreement No. 139-15-2025-013, dated June 20, 2025, IGK 000000C313925P4B0002). 

We thank Daniil Mangazeev, Pavel Kolesnik, and Ivan Sosin for their assistance in conducting experiments with the real robot.

\section*{Appendix}

This appendix provides additional analyses, implementation details, and experimental results that complement the main paper.

\begin{itemize}
    \item \cref{app:hyperparameters} describes the hyperparameters used for the policy model and algorithms.
    \item \cref{app:analysis} presents an analysis of the employed training strategies.
    \item \cref{sec:threshold_analysis} examines the role of the entropy threshold in the training strategies.
    \item \cref{sec:confidence_analysis} analyzes the confidence threshold of the segmentation model and its impact on the navigation performance of OVSegDT.
    \item \cref{app:final_exp} provides a breakdown and analysis of the individual training loss components.
    \item \cref{sec:statistical_analysis} presents a statistical analysis of the main experimental results.
    \item \cref{sec:observation_components} evaluates the impact of observation components on navigation performance.
    \item \cref{sec:segm_errs} analyzes the relationship between segmentation error and navigation quality.
    \item \cref{sec:performance_analysis} reports performance metrics, including inference speed and GPU memory usage.
    \item \cref{sec:real-world} presents results and insights from real-world robot experiments.
    \item \cref{sec:sup_related_works} provides an extended overview of related work on entropy-guided adaptive learning methods.
\end{itemize}

{
    \small
    \bibliographystyle{ieeenat_fullname}
    \bibliography{main}
}

\renewcommand{\thesection}{\Alph{section}}
\renewcommand{\thesubsection}{\thesection.\arabic{subsection}}
\renewcommand{\thefigure}{\Alph{figure}}
\renewcommand{\thetable}{\Alph{table}}
\setcounter{section}{0}
\setcounter{table}{0}
\setcounter{figure}{0}

\maketitlesupplementary

\section{Hyperparameters} \label{app:hyperparameters}

\cref{tab:model_hyperparameters} lists the key hyperparameters used for the policy model.

\begin{table}[htbp]
  \centering
  \scriptsize
  \caption{Policy Model Hyperparameters}
  \label{tab:model_hyperparameters}
   \begin{tabular}{lll}
    \hline
    \textbf{Component}     & \textbf{Parameter}                   & \textbf{Value}                          \\ \hline
\multirow{5}{*}{\parbox{1.2cm}{Visual encoder}}     
& \parbox{3.0cm}{Backbone Network}                   & SigLIP        \\ \cline{2-3}

& \parbox{3.0cm}{Fusion Type}                        & Concatenation \\ \cline{2-3}

& \parbox{3.0cm}{Use Visual Query}                   & True          \\ \cline{2-3}

& \raisebox{-2ex}{\rule{0pt}{5ex}}{\parbox{2.5cm}{Use Residual Connections}}           & True          \\

    \hline

\multirow{7}{*}{\parbox{1.7cm}{Transformer}} 
& \parbox{3.0cm}{Base Model}                & LLaMA \\ \cline{2-3}

& \parbox{3.0cm}{Number of Layers}          & 4     \\ \cline{2-3}

& \raisebox{-2ex}{\rule{0pt}{5ex}}\parbox{3.0cm}{Number of Attention Heads} & 8     \\ \cline{2-3}

& \raisebox{-2ex}{\rule{0pt}{5ex}}\parbox{3.0cm}{Hidden Dimension Size}     & 512   \\ \cline{2-3}

& \raisebox{-2ex}{\rule{0pt}{5ex}}\parbox{2.3cm}{MLP Hidden Dimension Size} & 1024  \\ \cline{2-3}

& \parbox{3.0cm}{Max Context Length}        & 100   \\ \cline{2-3}

& \parbox{3.0cm}{Shuffle Position IDs}      & True  \\
    \hline
    \multirow{2}{*}{\parbox{1.7cm}{Training parameters}} & {\parbox{3.0cm}{Learning Rate}}                 & 2.5 $\times$ 10$^{-4}$                 \\ \cline{2-3}
                                                & {\parbox{3.2cm}{Parallel environments}} & 40                                     \\ \hline

  \end{tabular}
\end{table}

\begin{table}[htbp]
  \centering
  \scriptsize
  \caption{Proximal Policy Optimization (PPO) Hyperparameters}
  \label{tab:ppo_hyperparameters}
  \begin{tabular}{lll}
    \hline
    \textbf{Component} & \textbf{Parameter}             & \textbf{Value}               \\ \hline
    \multirow{16}{*}{PPO}                & Clip Parameter ($\epsilon$)    & 0.2                          \\ \cline{2-3}
                        & PPO Epochs                     & 1                            \\ \cline{2-3}
                        & Mini-batches                   & 2                            \\ \cline{2-3}
                        & Value Loss Coefficient         & 0.5                          \\ \cline{2-3}
                        & Entropy Coefficient            & 0.01                         \\ \cline{2-3}
                        & Learning Rate (LR)             & 2.5 $\times$ 10$^{-4}$       \\ \cline{2-3}
                        & Adam Epsilon ($\epsilon$)      & 1 $\times$ 10$^{-5}$         \\ \cline{2-3}
                        & Max Gradient Norm              & 0.2                          \\ \cline{2-3}
                        & Steps per Update               & 100                          \\ \cline{2-3}
                        & Use GAE                        & True                         \\ \cline{2-3}
                        & Discount Factor ($\gamma$)     & 0.99                         \\ \cline{2-3}
                        & GAE Lambda ($\lambda$ / $\tau$) & 0.95                         \\ \cline{2-3}
                        & Linear Clip Decay              & False                        \\ \cline{2-3}
                        & Linear LR Decay                & True                         \\ \cline{2-3}
                        & Reward Window Size             & 50                           \\ \cline{2-3}
                        & Normalized Advantage           & False                        \\ \hline
  \end{tabular}
\end{table}

\begin{table}[htbp]
  \centering
  \scriptsize
  \caption{Agent and Observation Configuration Parameters}
  \label{tab:agent_obs_config}
  \begin{tabular}{lll}
    \hline
    \textbf{Component}        & \textbf{Parameter}            & \textbf{Value}              \\ \hline
    \multirow{2}{*}{NavMesh}  & Agent Max Climb               & 0.1                         \\ \cline{2-3}
                              & Cell Height                   & 0.05                        \\ \hline
    Environment               & Turn Angle (degrees)          & 30                          \\ \hline
    \multirow{5}{*}{Agent}    & Height                        & 1.41                        \\ \cline{2-3}
                              & Radius                        & 0.17                        \\ \cline{2-3}
                              & RGB Sensor Width (pixels)         & 360                         \\ \cline{2-3}
                              & RGB Sensor Height (pixels)        & 640                         \\ \cline{2-3}
                              & Horizontal FOV (degrees)      & 42                          \\ \cline{2-3}
                              & RGB Sensor Position (x, y, z)     & [0, 1.31, 0]                \\ \hline
  \end{tabular}
\end{table}

\cref{tab:ppo_hyperparameters} lists the key hyperparameters used for the Proximal Policy Optimization (PPO) algorithm.

\cref{tab:agent_obs_config} details the configuration parameters for the agent and its observations within the simulation environment.

\section{Analysis of Training Strategies} \label{app:analysis} 

\begin{figure*}[t]
    \centering
    \begin{minipage}{0.48\textwidth}
        \centering
        \includegraphics[width=\linewidth]{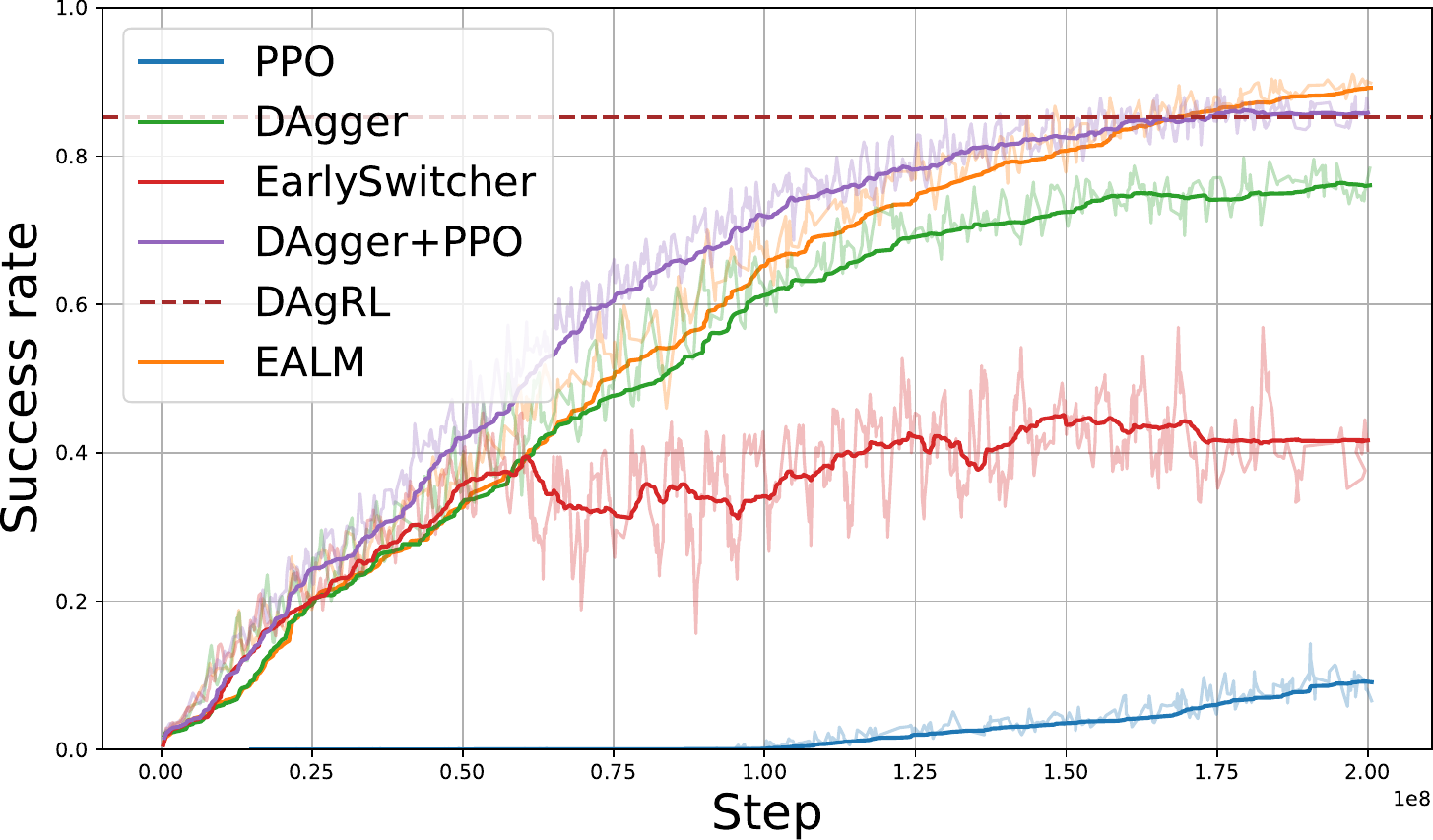}
        \caption{Training curves of \textbf{Success Rate} (higher is better) for the considered switching strategies. Our entropy-adaptive EALM reaches top performance with the fewest samples.}
        \label{fig:ovon_sr}
    \end{minipage}
    \hfill
    \begin{minipage}{0.48\textwidth}
        \centering
        \includegraphics[width=\linewidth]{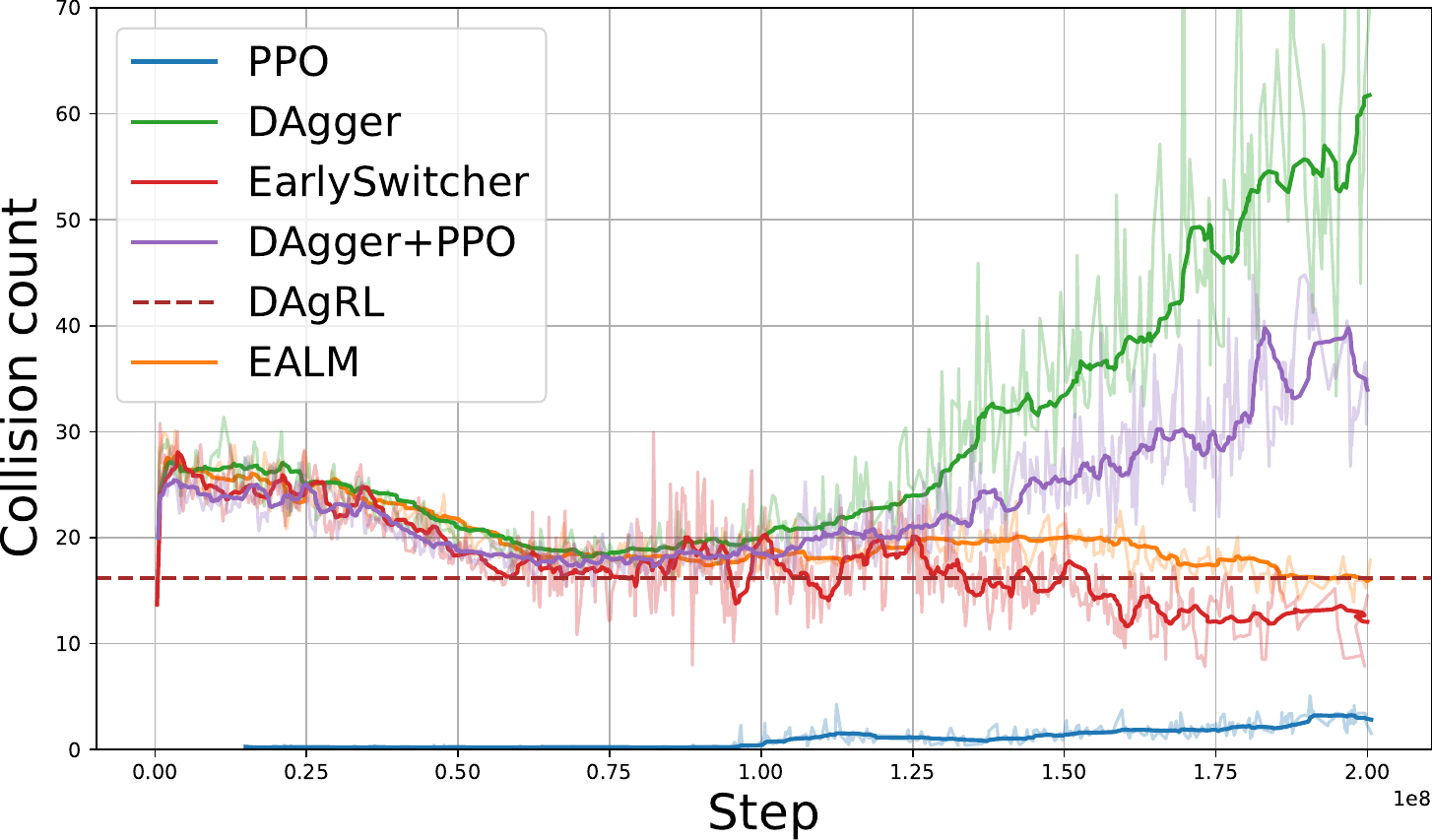}
        \caption{Training curves of \textbf{Collision Count} (lower is better). Collisions serve as a safety proxy; EALM steadily reduces collisions whereas other methods eventually over-fit and regress.}
        \label{fig:ovon_dg}
    \end{minipage}
\end{figure*}

As shown in ~\cref{fig:ovon_sr}, the naïve PPO agent makes little progress. However, Success Rate (SR) alone doesn't reveal \emph{how} the agent reaches its goal. To address this, ~\cref{fig:ovon_dg} includes the average number of \emph{collisions}, a useful proxy for safety since it's directly tied to the PPO reward—fewer collisions suggest the agent is genuinely learning to navigate rather than memorizing paths. While DAgger achieves near-perfect SR on training scenes, it averages over 50 collisions per episode, indicating strong overfitting. The DAgger+PPO combined loss reduces collisions initially, but because the PPO signal doesn't fully outweigh the DAgger loss, the agent regresses to unsafe behavior, leading to a drop in validation performance.

The EarlySwitcher highlights the sensitivity of manual objective scheduling. Gradually shifting from imitation to reinforcement between 40M and 60M steps 
reduces collisions (to about \mbox{26}), but it takes over 150M steps
for the policy to regain the SR lost during the transition. Identifying the optimal switching point would require extensive hyperparameter tuning.
EALM avoids this issue by \emph{continuously} adjusting the balance between imitation learning and reinforcement learning based on policy entropy. When the agent is uncertain, imitation dominates; as confidence grows, reinforcement naturally takes over. This adaptive approach combines the strengths of both methods—EALM converges twice as fast as the best baseline while achieving the fewest collisions.

\begin{figure*}[ht]
\centering
\begin{minipage}[t]{0.3\textwidth}
    \centering
    \includegraphics[width=\linewidth]{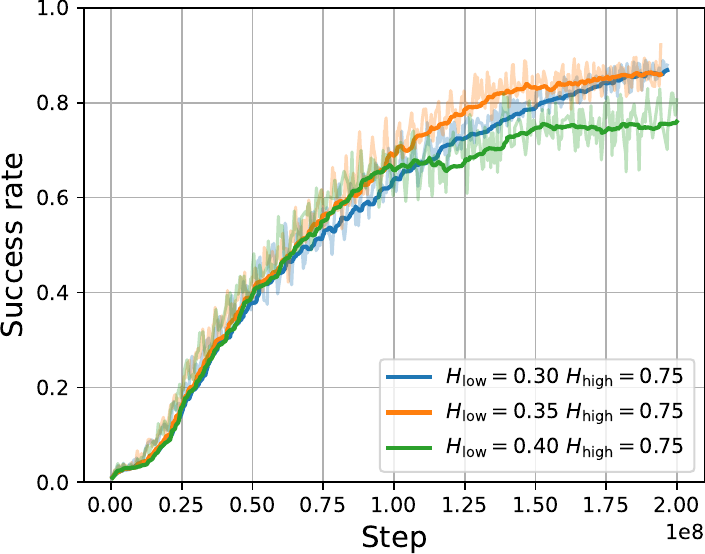}
    \caption{Success Rate}
    \label{fig:threshold_ablation_a}
\end{minipage}
\hfill
\begin{minipage}[t]{0.3\textwidth}
    \centering
    \includegraphics[width=\linewidth]{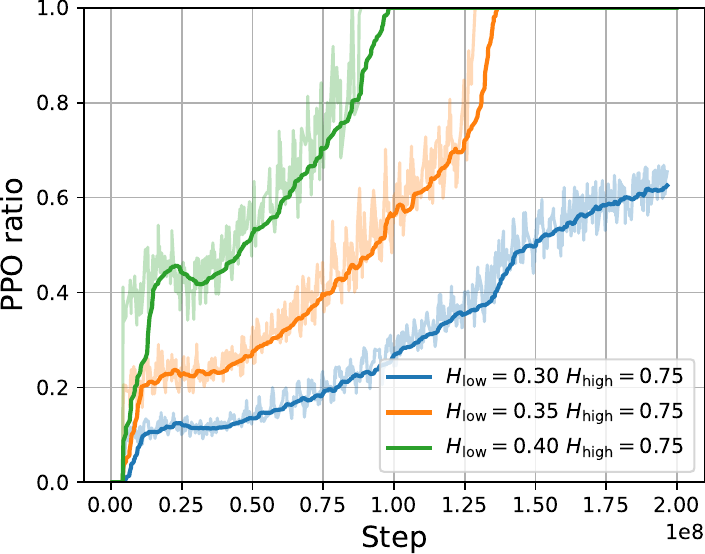}
    \caption{PPO Ratio}
    \label{fig:threshold_ablation_b}
\end{minipage}
\hfill
\begin{minipage}[t]{0.3\textwidth}
    \centering
    \includegraphics[width=\linewidth]{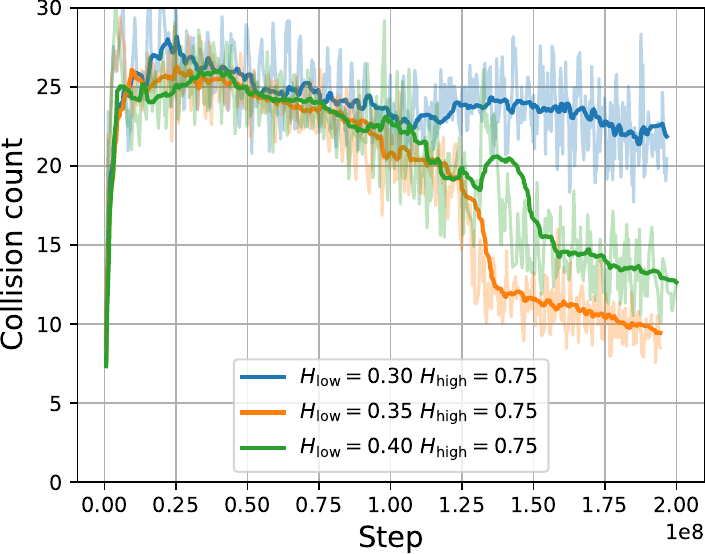}
    \caption{Collision Count}
    \label{fig:threshold_ablation_c}
\end{minipage}
\caption{Ablation study of entropy thresholds in EALM. We vary the lower entropy bound $H_{low}$ while keeping $H_{high}=0.75$ fixed. The $(H_{low}=0.35, H_{high}=0.75)$ configuration achieves optimal balance, reaching the highest success rate while maintaining the lowest collision count.}
\label{fig:threshold_ablation}
\end{figure*}

\begin{figure*}[t]
    \centering
    \begin{minipage}{0.32\textwidth}
        \centering
        \includegraphics[width=\linewidth]{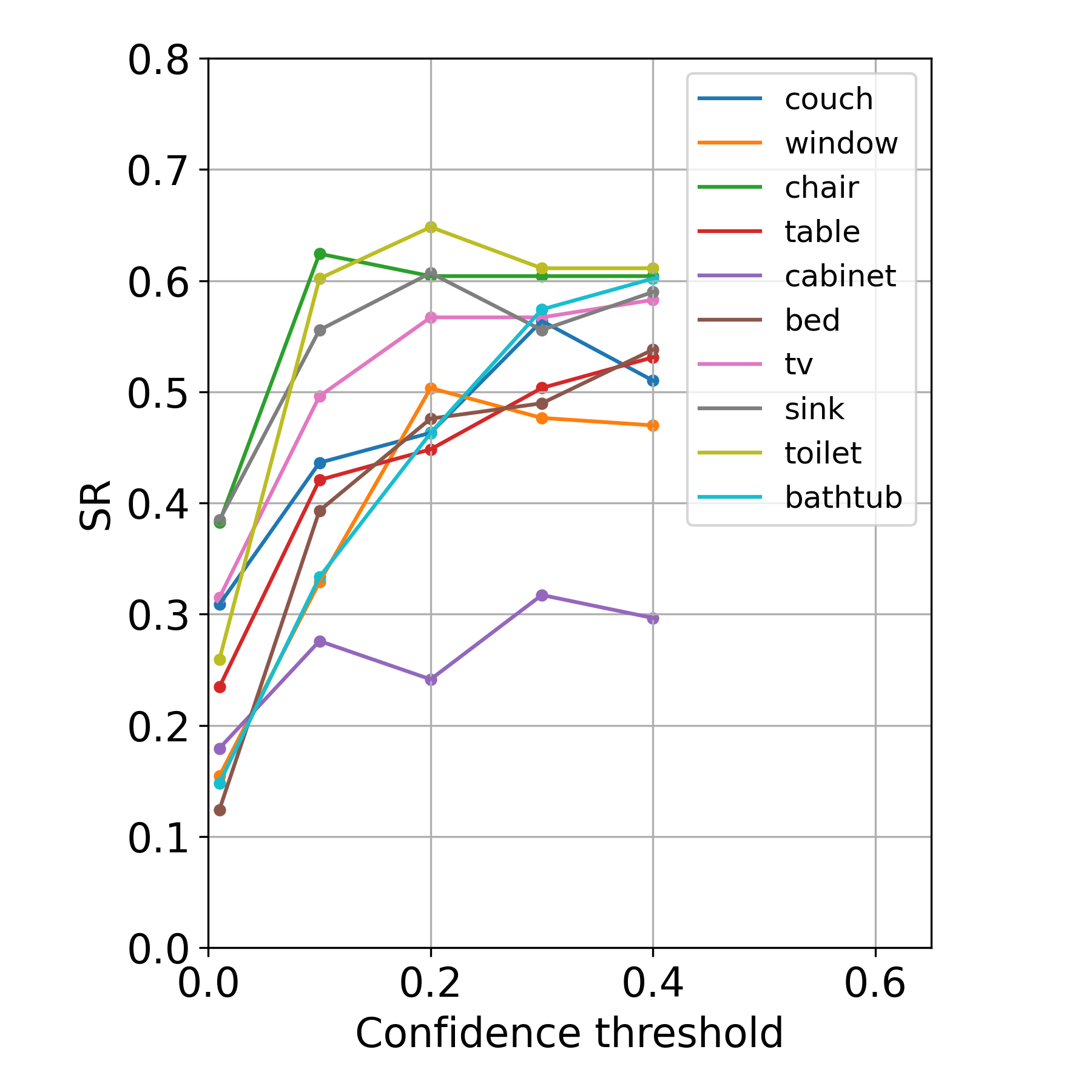}
        \caption{Dependence of the success rate (SR) on the confidence threshold of the YOLOE model for different categories from the \textit{val seen} split of HM3D-OVON. The plots are shown for the top 10 categories by number of episodes. }
        \label{fig:ovon_val_seen}
    \end{minipage}
    \hfill
    \begin{minipage}{0.32\textwidth}
        \centering
        \includegraphics[width=\linewidth]{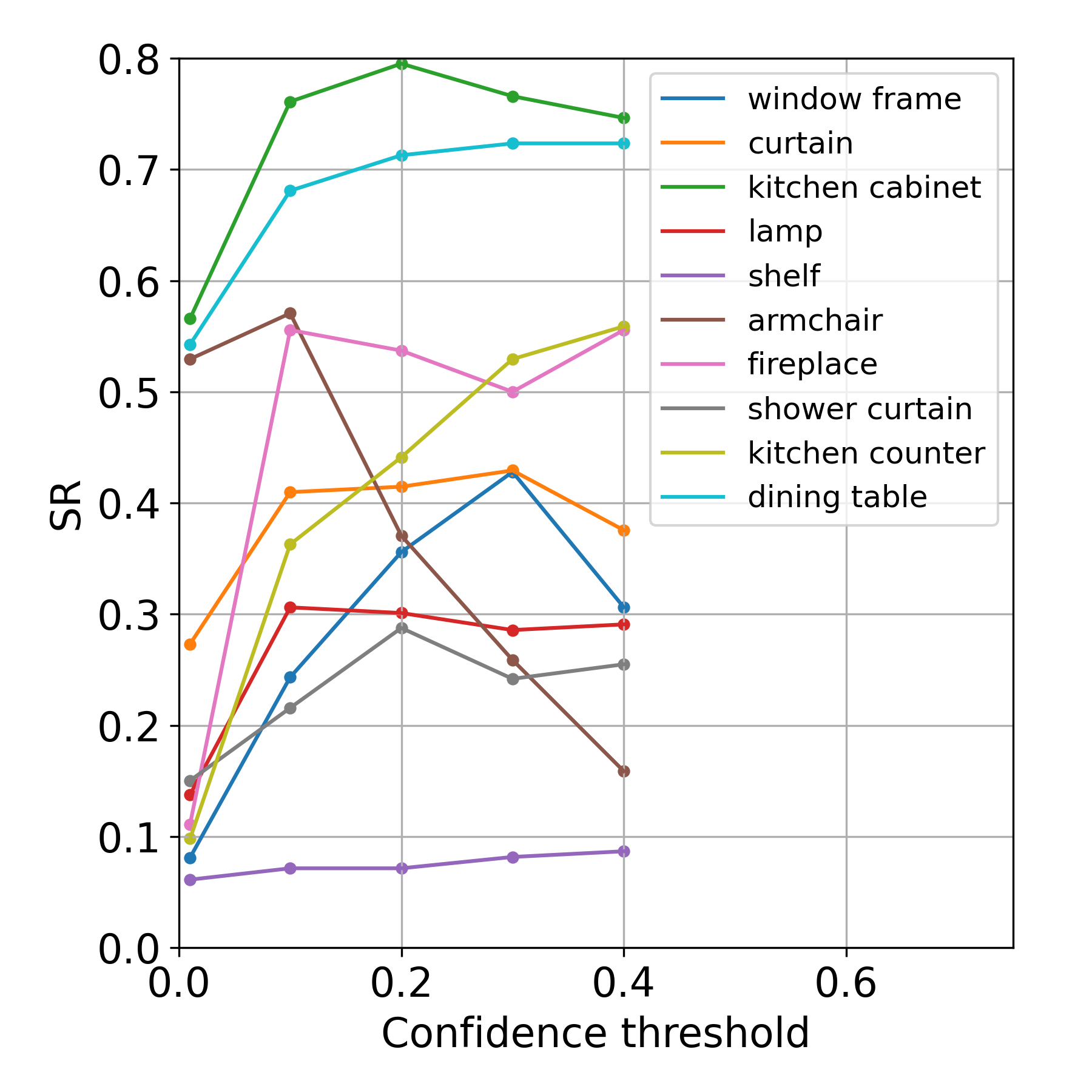}
        \caption{Dependence of the success rate (SR) on the confidence threshold of the YOLOE model for different categories from the \textit{val seen synonyms} split of HM3D-OVON. The plots are shown for the top 10 categories by number of episodes.}
        \label{fig:ovon_val_seen_synonyms}
    \end{minipage}
    \hfill
    \begin{minipage}{0.32\textwidth}
        \centering
        \includegraphics[width=\linewidth]{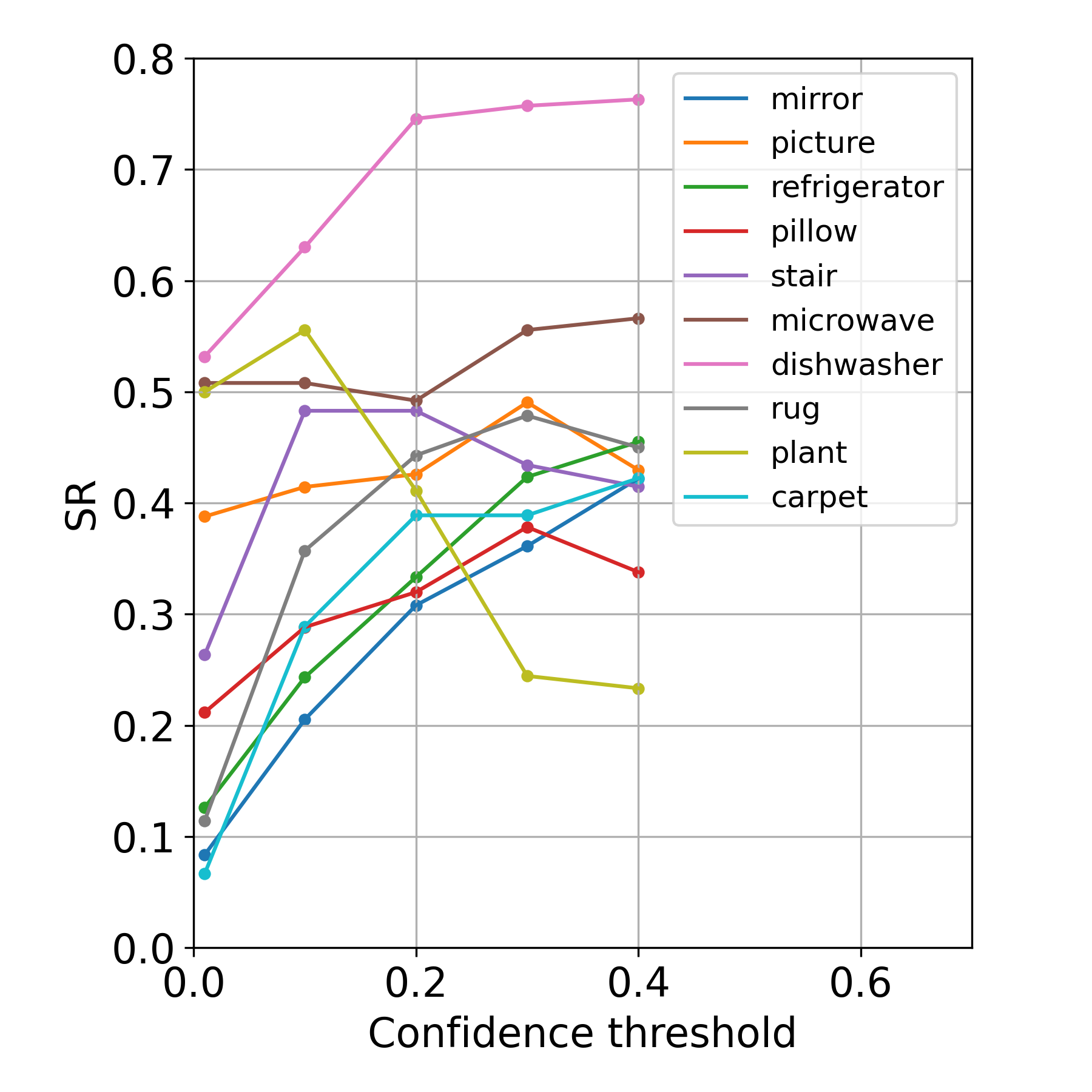}
        \caption{Dependence of the success rate (SR) on the confidence threshold of the YOLOE model for different categories from the \textit{val unseen} split of HM3D-OVON. The plots are shown for the top 10 categories by number of episodes.}
        \label{fig:ovon_val_unseen}
    \end{minipage}
\end{figure*}

\section{Analysis of the Entropy Threshold of the Training Strategies}
\label{sec:threshold_analysis}

\paragraph{Objective.}
The efficacy of EALM relies on defining an appropriate "confidence" window for the policy. The hyperparameters $H_{low}$ and $H_{high}$ determine the entropy range over which the agent transitions from Imitation Learning (IL) to Reinforcement Learning (RL). This section analyzes how shifting this window—specifically the lower bound $H_{low}$—affects training dynamics and final performance.

\paragraph{Experimental Setup.}
We compare three configurations for the lower entropy bound: $H_{low} \in \{0.30, 0.35, 0.40\}$, while keeping the upper bound fixed at $H_{high}=0.75$. 
The upper bound $H_{high}$ represents the entropy level at which the policy begins to accept partial RL signals.
The lower bound $H_{low}$ represents the point of maximum confidence where the IL signal is fully deactivated ($\lambda_{PPO} = 1$).

\paragraph{Dynamics Analysis.}
~\cref{fig:threshold_ablation} illustrates the impact of these thresholds on three key metrics:
\begin{itemize}
    \item \textbf{PPO Ratio (Transition Speed).} The center plot (\cref{fig:threshold_ablation_b}) shows the evolution of the mixing coefficient $\lambda_t$ (PPO weight) over time. A lower $H_{low}$ (e.g., $0.30$, blue curve) requires the agent to achieve higher certainty (lower entropy) before fully switching to PPO. Consequently, the transition to pure RL is slower compared to the $0.40$ setting (green curve), which switches earliest.
    
    \item \textbf{Success Rate (Performance).} The left plot (\cref{fig:threshold_ablation_a}) reveals that switching too early ($H_{low}=0.40$) is detrimental. The green curve exhibits a dip in success rate mid-training because the agent stops imitating the expert before it has fully mastered the necessary navigation primitives. Conversely, switching too late ($H_{low}=0.30$, blue curve) delays the benefits of RL exploration, slowing down the mastery of complex scenarios.
    
    \item \textbf{Collision Count (Safety).} The right plot (\cref{fig:threshold_ablation_c}) demonstrates the safety trade-off. Prolonged imitation ($H_{low}=0.30$) leads to higher collision rates in the intermediate phase because the agent overfits to expert trajectories without receiving negative RL rewards for collisions. The balanced configuration ($H_{low}=0.35$, orange curve) achieves the optimal compromise, minimizing collisions while maximizing success.
\end{itemize}

\paragraph{Conclusion.}
The configuration $(H_{low}=0.35, H_{high}=0.75)$ yields the best performance. This empirically validates our heuristic derived in the main text, balancing the need for stabilizing demonstrations with the necessity of safety-critical reinforcement learning.

\section{Analysis of the Confidence Threshold of the Segmentation Model on Navigation Performance of OVSegDT}
\label{sec:confidence_analysis}

In our experiments with predicted segmentation, we use a pretrained YOLOE model based on YOLOv8-L. We analyze how the confidence threshold of the YOLOE model for each category affects the percentage of episodes with successful navigation to objects of that category. ~\cref{fig:ovon_val_seen}, \cref{fig:ovon_val_seen_synonyms}, and \cref{fig:ovon_val_unseen} illustrate the relationship between success rate and different confidence thresholds for the top 10 categories by number of episodes from each split of HM3D-OVON.

\begin{figure*}[ht] 
\centering
    \centering
    \includegraphics[width=\linewidth]{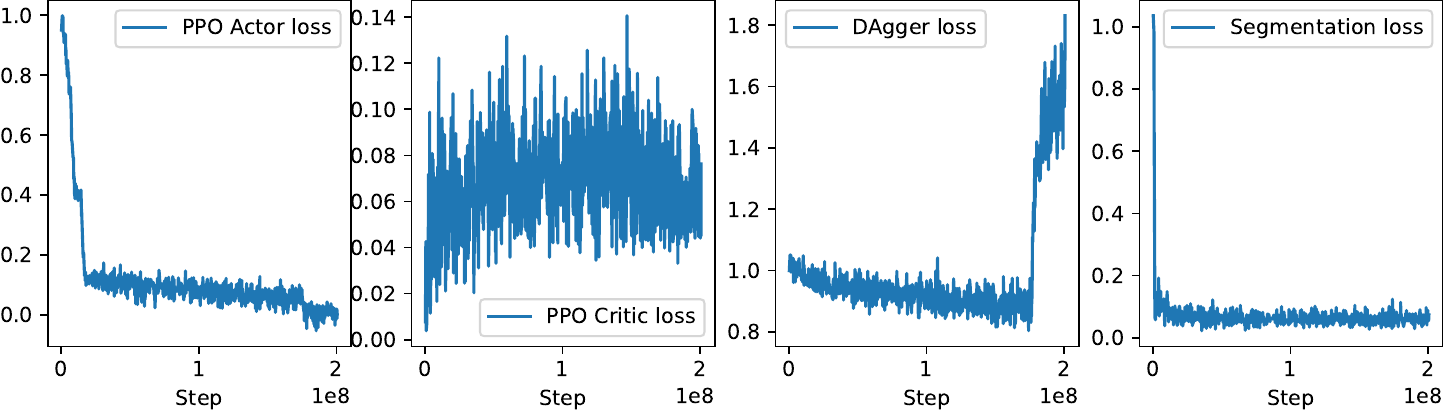}
    \caption{Training losses for OVSegDT over 200\,M environment steps.
\textbf{Left to right:}
(\emph{i}) PPO actor loss;
(\emph{ii}) PPO critic loss;
(\emph{iii}) DAgger (behavior-cloning) loss;
(\emph{iv}) Segmentation loss.}
    \label{fig:Final_losses}
\end{figure*}
\begin{figure*}[ht] 
\centering
    \centering
    \includegraphics[width=\linewidth]{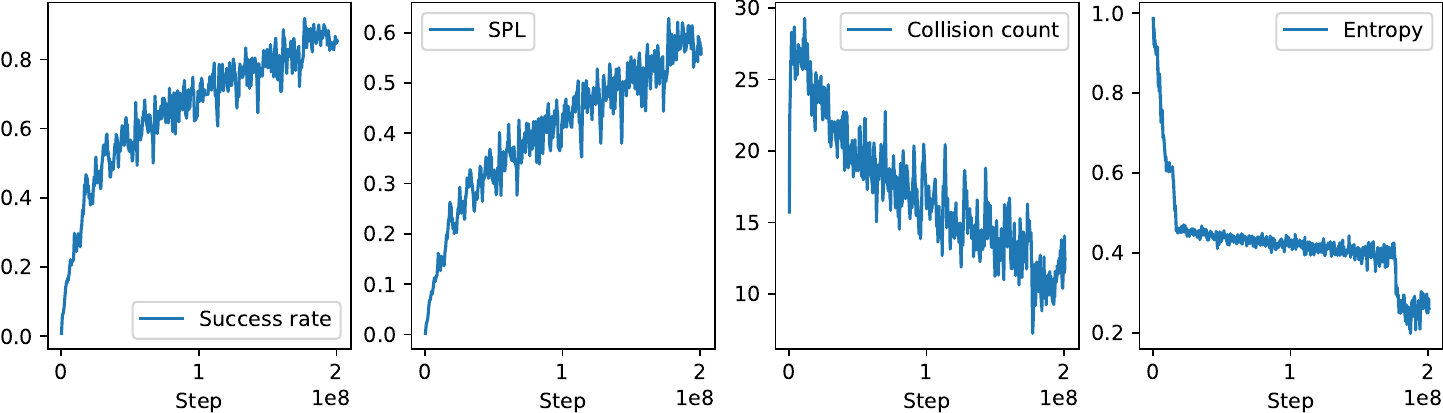}
    \caption{Metrics on the HM3D-OVON \textit{train} split during training.
\textbf{Left to right:}
(\emph{i}) Success Rate (SR);
(\emph{ii}) Success weighted by Path Length (SPL);
(\emph{iii}) Collision count;
(\emph{iv}) Policy entropy.}
    \label{fig:Final_metrics}
\end{figure*}

The analysis shows that some categories are challenging for the pretrained YOLOE model and require a low confidence threshold for successful recognition (e.g., \textit{plant} from \textit{val unseen} or \textit{armchair} from\textit{ val seen synonyms}). On the other hand, some categories require a high confidence threshold to be reliably distinguished from others (e.g.,\textit{ kitchen counter} from \textit{val seen synonyms} or mirror from \textit{val unseen}). For most categories, the optimal confidence threshold falls within the range of 0.1 to 0.3.

\section{Analysis of Training Loss Components} \label{app:final_exp} 

~\cref{fig:Final_losses} visualizes how each loss term evolves over 200\,M environment steps.  
The \emph{PPO actor} loss drops within the first 10\,M steps and then asymptotically approaches zero, reflecting rapid policy improvement.  
The \emph{critic} loss spikes early, stabilizing once value estimates become consistent.  
During the imitation-heavy phase the \emph{DAgger} loss is low; it rises after $\sim$160\,M steps when EALM hands full control to PPO, confirming that the optimization objective has indeed shifted.  
Finally, the \emph{segmentation} loss falls quickly and remains an order of magnitude lower than the policy losses, indicating that the auxiliary head is able to reconstruct accurate goal masks throughout training.

\begin{table*}[h!]
\centering
\caption{Statistical comparison of different switching strategies on HM3D-OVON benchmark.}
\scriptsize
\begin{tabular}{llllp{3.2cm}}
\hline
\textbf{H$_0$ Hypothesis} & \textbf{Metric} & \textbf{Split} & \textbf{p-value} & \textbf{Conclusion} \\
\hline
No significant evidence that DAgger+PPO outperforms DAgger. & collisions & val seen & 0.004 & Significant difference \\
No significant evidence that DAgger+PPO outperforms DAgger. & collisions & val unseen & 0.03 & Significant difference \\
\hline
No significant evidence that EarlySwitcher outperforms DAgger+PPO. & collisions & val seen & 0.008 & Significant difference \\
No significant evidence that EarlySwitcher outperforms DAgger+PPO. & collisions & val unseen & 0.01 & Significant difference \\
\hline
No significant evidence that EALM outperforms DagRL. & SR & val seen & 0.001 & Significant difference \\
No significant evidence that EarlySwitcher outperforms DAgger+PPO. & SR & val unseen & 0.03 & Significant difference \\
\hline
No significant evidence that EarlySwitcher outperforms DAgger+PPO. & SPL & val unseen & 0.002 & Significant difference \\
\hline

\end{tabular}

\label{tab:stat-comparison-training}
\end{table*}

\begin{table*}[h!]
\centering
\caption{Statistical comparison of different observation types and training objectives on Val Unseen split of HM3D-OVON benchmark.}
\scriptsize
\begin{tabular}{llllp{2.3cm}}
\hline
\textbf{H$_0$ Hypothesis} & \textbf{Metric} & \textbf{Split} & \textbf{p-value} & \textbf{Conclusion} \\
\hline
No significant evidence that OVSegDT+Goal Mask+\textbf{$r^{\text{sem}}$}+\textbf{$\mathcal{L}_{\text{seg}}$} outperforms OVSegDT+Goal Mask+\textbf{$r^{\text{sem}}$}. & SR & val unseen & 0.0003 & Significant difference \\
No significant evidence that OVSegDT+Goal Mask+\textbf{$r^{\text{sem}}$} +\textbf{$\mathcal{L}_{\text{seg}}$} outperforms OVSegDT+Goal Mask+\textbf{$r^{\text{sem}}$}. & SPL & val unseen & 0.0001 & Significant difference \\
\hline
No significant evidence that OVSegDT+Goal Mask+\textbf{$r^{\text{sem}}$}+\textbf{$\mathcal{L}_{\text{seg}}$} outperforms OVSegDT+Goal Mask+\textbf{$\mathcal{L}_{\text{seg}}$}. & SR & val unseen & $10^{-6}$ & Significant difference \\
No significant evidence that OVSegDT+Goal Mask+\textbf{$r^{\text{sem}}$} +\textbf{$\mathcal{L}_{\text{seg}}$} outperforms OVSegDT+Goal Mask+\textbf{$\mathcal{L}_{\text{seg}}$}. & SPL & val unseen & $10^{-6}$ & Significant difference \\
\hline
No significant evidence that OVSegDT+Goal Mask+\textbf{$r^{\text{sem}}$}+\textbf{$\mathcal{L}_{\text{seg}}$} outperforms OVSegDT. & collisions & val unseen & 0.02 & Significant difference \\
\hline

\end{tabular}

\label{tab:stat-comparison-observations}
\end{table*}

\begin{table*}[h!]
\centering
\caption{Statistical comparison of different strategies of adaptation to predicted segmentation for OVSegDT method on Val Unseen split of HM3D-OVON benchmark.}
\scriptsize
\begin{tabular}{p{10cm}lllp{2.3cm}}
\hline
\textbf{H$_0$ Hypothesis} & \textbf{Metric} & \textbf{Split} & \textbf{p-value} & \textbf{Conclusion} \\
\hline
No significant evidence that OVSegDT+YOLOE calibrated outperforms OVSegDT+YOLOE. & SR & val unseen & $3\cdot10^{-5}$ & Significant difference \\

No significant evidence that OVSegDT+YOLOE calibrated outperforms OVSegDT+YOLOE. & SPL & val unseen & 0.0006 & Significant difference \\
\hline
No significant evidence that OVSegDT+YOLOE calibrated+Predicted mask finetune outperforms OVSegDT+YOLOE calibrated. & SR & val unseen & $10^{-5}$ & Significant difference \\

No significant evidence that OVSegDT+YOLOE calibrated+Predicted mask finetune outperforms OVSegDT+YOLOE calibrated. & SPL & val unseen & $10^{-4}$ & Significant difference \\
\hline
\end{tabular}

\label{tab:stat-comparison-segmentation}
\end{table*}

Performance trends on the HM3D-OVON \textit{train} split are shown in ~\cref{fig:Final_metrics}.  
Both \textbf{Success Rate} and \textbf{SPL} increase monotonically, while the \textbf{collision count} falls to fewer than ten per episode - evidence that the agent learns safer, shorter paths as training progresses.  
The \textbf{entropy} curve drops sharply during the first imitation-dominated stage, slowly decreases while EALM mixes objectives, and exhibits a second rough decline when the algorithm switches to a PPO-only loss near the end of training.  
Importantly, this late entropy reduction \emph{does not} introduce instability: all task metrics continue to improve smoothly, demonstrating that EALM’s automatic transition preserves training stability and leads to a confident yet robust final policy.

\section{Statistical Analysis} \label{sec:statistical_analysis}

We analyze the statistical significance of the impact of different components of our method. Each algorithm is run three times, and we perform the one-sided t-test for unpaired samples to determine the level of statistical significance. ~\cref{tab:stat-comparison-training}, ~\cref{tab:stat-comparison-observations}, and ~\cref{tab:stat-comparison-segmentation} present the comparison results and corresponding conclusions. Thus, the conclusions presented in the main text are supported by the statistical significance of the results.

\section{Analysis of Observation Components Impact on Navigation Performance} 
\label{sec:observation_components}

The OVSegDT method uses two types of goal prompts at inference time: a textual instruction and a visual cue in the form of a binary mask of the target object. We analyze the impact of each of these components during inference in ~\cref{tab:ablation-input}. The observation setting Goal Mask+Text Goal corresponds to the original version of OVSegDT. When using only the Goal Mask at inference time, we replace the part of the observation corresponding to the text instruction with a dummy zero input. In this case, we observe a drop in navigation performance. In the case of navigation using only the text goal, we always feed a zero mask to the model during inference. This leads to a complete breakdown of OVSegDT’s navigation, highlighting the importance of high-quality masks for successful navigation. Thus, although binary goal masks are a key component of the observation, the textual instruction also contributes to improving navigation quality in OVSegDT.

\begin{table}[h!]
\centering
\caption{Analysis of the impact of the goal mask and textual instruction on navigation performance on Val Unseen of HM3D-OVON benchmark.}
\scriptsize
\begin{tabular}{@{}p{0.7cm}p{2.2cm}p{1.2cm}p{1.4cm}p{1.1cm}@{}}
\hline
\textbf{Method} & \textbf{Observation} & \textbf{SR} & \textbf{SPL} & \textbf{Collisions} \\
\hline
OVSegDT & Goal Mask+Text Goal & $\mathbf{44.7} \pm 0.4 $ & $\mathbf{20.6} \pm 0.2$ & $\underline{45.4} \pm 0.6 $  \\
OVSegDT & Goal Mask  & $\underline{12.6} \pm 0.6 $ & $\underline{4.5} \pm 0.2 $ & $\underline{70.0} \pm 0.5 $ \\
OVSegDT & Text Goal & $0.1 \pm 0.07$ & $0.01 \pm 0.02 $ & $68.0 \pm 0.2$ \\
\hline
\end{tabular}

\label{tab:ablation-input}
\end{table}

\section{Analysis of Segmentation Error and Navigation Quality}
\label{sec:segm_errs}
We analyze how the adaptation methods we introduced for applying OVSegDT to the predicted segmentation affect segmentation and navigation quality. To evaluate segmentation quality, we use the false positive rate (FPR) and false negative rate (FNR) metrics. A detailed definition of these metrics can be found in ~\cref{sec:segm-evaluation-metrics}. As shown in ~\cref{tab:segm_analysis}, the main source of errors is false negative predictions. Introducing confidence-threshold calibration for YOLOE reduces the FPR and thus improves navigation quality. Fine-tuning on predicted masks improves navigation quality by reducing the false negative rate, i.e. the model learns to find trajectories in which YOLOE can successfully segment the target objects. Finally, we assess the robustness of the confidence-calibration process to photometric augmentations. A detailed description of the augmentations used is provided in ~\cref{sec:augm}.
Owing to the pre-trained YOLOE, the segmentation performance remains stable under these perturbations (see ~\cref{tab:segm_analysis}). Calibration is introduced to mitigate errors arising from synonym overlap within the navigation vocabulary and to address the real-to-simulation domain gap, particularly for classes such as plant, stairs, curtains.


\begin{table}[h!]
\centering
\caption{Analysis of the segmentation errors source and the navigation performance on Val Unseen of HM3D-OVON benchmark.}
\scriptsize
\begin{tabular}{@{}p{1.5cm}p{1.6cm}@{}p{1.15cm}p{1.15cm}p{1.15cm}}
\hline
\textbf{Segmentation method} & \textbf{Predicted mask fine-tune} & \textbf{SR, \%} & \textbf{FPR, \%} & \textbf{FNR, \%} \\
\hline
YOLOE             & \xmark & $28.8 \pm 0.4 $ & $11.9 \pm 0.3 $ & $68.0 \pm 0.3 $ \\
YOLOE calib.  & \xmark  & $35.7 \pm 0.4 $  & $\mathbf{5.5} \pm 0.3 $ & $68.5 \pm 0.3 $ \\
YOLOE calib.  & \cmark & $\mathbf{44.7} \pm 0.4 $  & $\underline{7.2} \pm 0.2 $ & $\mathbf{61.3} \pm 0.3 $ \\
YOLOE calib. (image aug.) & \cmark & $\underline{42.2} \pm 0.7 $  & $7.3 \pm 0.2 $ & $\underline{63.6} \pm 0.3 $ \\
\hline
\end{tabular}
\label{tab:segm_analysis}
\end{table}

\subsection{Random Lighting Augmentation}
\label{sec:augm}
We apply a random lighting augmentation module to each input image
$\mathbf{I} \in [0,1]^{C \times H \times W}$ to evaluate
navigation robustness under controlled photometric perturbations.
The augmentation operates per-image and uses the hyperparameters listed in ~\cref{tab:augmentation}.

\begin{table}[h]
\centering
\scriptsize
\caption{Hyperparameters used in the Random Lighting Augmentation.}
\begin{tabular}{l c}
\toprule
\textbf{Parameter} & \textbf{Value} \\
\midrule
$p_\text{bright}$ & 0.5 \\
$p_\text{contrast}$ & 0.5 \\
$p_\text{satur}$ & 0.3 \\
$p_\text{hue}$ & 0.1 \\
$\gamma^{dist}_{\min}$ & 0.6 \\
$\gamma^{dist}_{\max}$ & 1.4 \\
$p_{\text{shadow}}$ & 0.5 \\
$s_\text{min}$ & 0.4 \\
$s_\text{max}$ & 0.8 \\
\bottomrule
\end{tabular}
\label{tab:augmentation}
\end{table}

\textbf{Color Jitter.}
Brightness, contrast, saturation, and hue are randomly perturbed independently.
For brightness, contrast, and saturation we sample
\begin{equation}
\alpha_p \sim \mathcal{U}(1-p,\; 1+p), \quad
p \in \{p_\text{bright}, p_\text{contrast}, p_\text{satur}\}.
\end{equation}
For hue we sample
\begin{equation}
\delta_h \sim \mathcal{U}(-p_\text{hue}, p_\text{hue}).
\end{equation}

The sequential jitter transformation applied to each image first adjusts its brightness, then adjusts its contrast, followed by a saturation adjustment, and finally applies a hue shift. Each of these operations uses independently sampled random factors for brightness, contrast, saturation, and hue.

\textbf{Gamma Correction.}
To introduce nonlinear photometric distortions, we sample
\begin{equation}
\gamma^{dist} \sim \mathcal{U}(\gamma^{dist}_{\min}, \gamma^{dist}_{\max}),
\end{equation}
and adjust each image by raising its pixel values to this exponent.

\textbf{Random Shadow.}
With probability $p_{\text{shadow}}$, a rectangular region
$R = [x_1,x_2]\times[y_1,y_2]$ is darkened by a multiplicative factor
\begin{equation}
s \sim \mathcal{U}(s_\text{min},\, s_\text{max}).
\end{equation}
This yields:
\[
\mathbf{I}(x,y) \leftarrow
\begin{cases}
s \cdot \mathbf{I}(x,y), & (x,y) \in R, \\
\mathbf{I}(x,y), & \text{otherwise}.
\end{cases}
\]

\textbf{Output.}
All transformed images are clipped to $[0,1]$ and reassembled into a batch.
This augmentation introduces illumination changes, contrast shifts,
nonlinear intensity warping, and localized shadows, enabling systematic
stress-testing of the segmentation model and therefore OVSegDT navigation under diverse lighting conditions.

\subsection{Segmentation Evaluation Metrics}
\label{sec:segm-evaluation-metrics}
To evaluate segmentation performance, we compute the False Positive Rate (FPR) and False Negative Rate (FNR) based on the pixel-wise predictions of the model. For each predicted mask $\text{pred\_mask}$ and its corresponding ground-truth mask $\text{gt\_mask}$, we first check if both masks are empty (i.e., no foreground pixels). In this case, we count it as a True Negative (TN). Otherwise, we compute the Intersection over Union (IoU) between the predicted and ground-truth masks. If the IoU exceeds $0.5$, the prediction is considered a True Positive (TP). If the IoU is below $0.5$ and the predicted mask has foreground pixels, it is counted as a False Positive (FP); if the predicted mask is empty while the ground-truth mask is not, it is counted as a False Negative (FN). Using these counts, FPR and FNR are computed as:

\begin{equation}  
\text{FPR} = \frac{\text{FP}}{\text{FP} + \text{TN}},  
\end{equation}  

\begin{equation}  
\text{FNR} = \frac{\text{FN}}{\text{FN} + \text{TP}}.  
\end{equation}  

These metrics provide a quantitative measure of over-segmentation (FPR) and under-segmentation (FNR) errors in the YOLOE model during the navigation process.

\begin{figure*}[ht]
\begin{center}
\centerline{\includegraphics[width=1\linewidth]{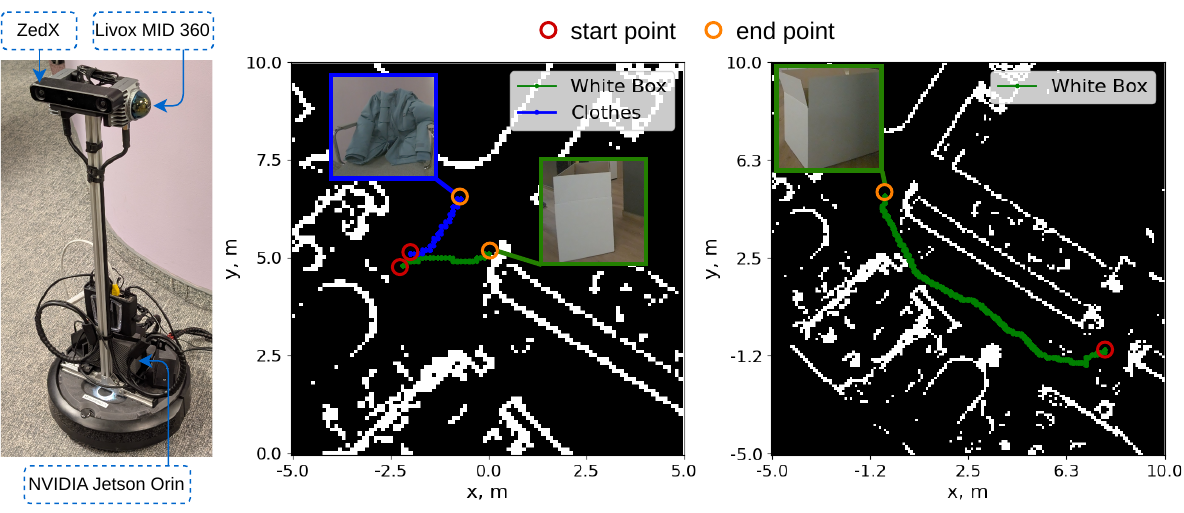}}
\caption{\textbf{Left:} iRobot Create 3 equipped with a ZED X camera, Livox MID 360 LiDAR, and Nvidia Jetson Orin PC.
\textbf{Right:} demonstration of the robot’s trajectories on the occupancy grid for two scenarios. In the first scenario, the robot must sequentially reach each of two goals specified in natural language. In the second scenario, the robot must find a target object that is outside its field of view while avoiding obstacles. }
\label{fig:real-robot}
\end{center}
\end{figure*}
\section{Performance Analysis}
\label{sec:performance_analysis}


\begin{table}[h!]
\centering
\scriptsize
\caption{Analysis of GPU memory usage and inference time of OVSegDT with predicted segmentation.}
\begin{tabular}{@{}p{4.6cm}p{1.2cm}p{1.2cm}@{}}
\hline
\textbf{Segmentation source} & \textbf{GT} & \textbf{YOLOE} \\
\hline
GPU Memory Usage (Gb) & $3486$ & $4992$ \\
Inference time (ms, NVIDIA H100) & $13.6 \pm 0.9$ & $65.0 \pm 14$ \\
Inference time (ms, NVIDIA Jetson Orin) & $107 \pm 5$ & $132 \pm 5$ \\
\hline
\end{tabular}
\label{tab:model_performance}
\end{table}

The performance analysis demonstrates that OVSegDT remains efficient across both ground-truth (GT) and predicted segmentation settings (see ~\cref{tab:model_performance}). While the use of YOLOE-based predicted segmentation increases GPU memory usage (from 3486 Gb to 4992 Gb) and inference latency on the NVIDIA H100 (from $13.6 \pm 0.9$ ms to $65.0 \pm 14$ ms), the impact on the embedded NVIDIA Jetson Orin platform is relatively modest (from $107 \pm 5$ ms to $132 \pm 5$ ms). This indicates that incorporating predicted segmentation does not significantly slow down the model in onboard scenarios, which are critical for deployment. Overall, OVSegDT demonstrates strong efficiency in terms of GPU memory usage and maintains near real-time inference performance, making it suitable for real-time navigation policy prediction even when relying on predicted segmentation inputs.

\section{Real-World Robot Experiments}
\label{sec:real-world}

We conduct demonstration experiments on a real iRobot Create 3 robot equipped with a ZED X camera, Livox MID 360 LiDAR, and an Nvidia Jetson Orin PC. Since the ZED X has a wide FoV (105\textdegree vs. 42\textdegree in the HM3D-OVON benchmark), we train a version of OVSegDT with this camera setup to improve the estimation of distances to objects and obstacles during real-robot experiments. ~\cref{fig:real-robot} demonstrates the effectiveness of the OVSegDT strategy for open-vocabulary navigation in two scenarios.

In the first scenario, the robot must distinguish between two targets in front of it: a white box and clothes. The robot’s trajectories show that it successfully accomplishes this task and, after detecting the target, moves directly toward it. In the second scenario, the robot must find a white box initially hidden from its field of view at the start of the episode. It also successfully completes this task. Additionally, when moving along long trajectories, the robot effectively avoids obstacles.

However, it is worth noting that during training, the robot receives a large reward for stopping within a small radius of the target (25 cm), while the penalty for collisions is relatively small. As a result, in real-world experiments, we observe risky behavior near the goal object. Balancing precise goal reaching with safe stopping in front of the target without a depth camera remains a subject for future work.

\section{Related Works on Entropy-Guided Adaptive Learning Methods} \label{sec:sup_related_works}

In this section, we provide an extended overview of related work on entropy-guided adaptive learning methods.

Recent advances in adaptive learning have increasingly leveraged entropy as a guiding signal. 
EntAugment~\cite{yang2024entaugment} dynamically adjusts data augmentation intensity based on model entropy to improve generalization, while EA-KD~\cite{su2025ea} utilizes entropy to re-weight samples during knowledge distillation, prioritizing high-uncertainty instances. 
In curriculum learning, the READ-C framework~\cite{satici2025autonomous} utilizes relative entropy for autonomous curriculum design, and in the graph domain, entropy has been used to guide curriculum learning for contrastive tasks~\cite{zeng2024multi}. 
REALM~\cite{seto2024realm} employs robust loss functions scaled by entropy to stabilize test-time adaptation. While our EALM similarly leverages entropy as an adaptive signal, it focuses on dynamically balancing imitation and reinforcement learning objectives during training rather than test-time robustness or curriculum sequencing.

In the realm of policy learning, SAC~\cite{haarnoja2018soft} maximizes policy entropy to encourage exploration. This contrasts with EALM, which uses entropy not as an exploration bonus, but as a scheduling signal to shift between learning paradigms.
SafeDAgger~\cite{zhang2016query} introduces a safety policy to reduce queries to expensive reference policies. EALM achieves a similar goal of efficiency but does so through automatic phase transitions guided by policy entropy, eliminating the need for an additional safety network.

Fundamentally, while these approaches use entropy to modulate hyperparameters or sample weights within a single paradigm, EALM employs policy entropy as a switching mechanism between two distinct objectives: imitation learning and reinforcement learning. EALM automatically orchestrates the transition from supervised bootstrapping to autonomous exploration, eliminating the need for manual phase scheduling.



\end{document}